\documentclass[10pt,twocolumn]{article}

\usepackage[margin=0.68in,top=0.72in,bottom=0.72in,columnsep=0.24in]{geometry}
\usepackage{graphicx}
\usepackage{microtype}
\usepackage{mathtools}
\usepackage{amssymb}
\usepackage{booktabs}
\usepackage{siunitx}
\usepackage{algorithm}
\usepackage{algpseudocode}
\usepackage{caption}
\usepackage[title]{appendix}
\usepackage{amsmath,amsfonts,bm}

\def\eqref#1{equation~\ref{#1}}
\def\1{\bm{1}}

\def\vtheta{{\bm{\theta}}}

\def\vv{{\bm{v}}}
\def\vw{{\bm{w}}}
\def\vx{{\bm{x}}}

\def\mI{{\bm{I}}}

\DeclareMathAlphabet{\mathsfit}{\encodingdefault}{\sfdefault}{m}{sl}
\SetMathAlphabet{\mathsfit}{bold}{\encodingdefault}{\sfdefault}{bx}{n}

\newcommand{\E}{\mathbb{E}}

\newcommand{\R}{\mathbb{R}}

\usepackage[authoryear,round]{natbib}
\usepackage[hidelinks]{hyperref}
\usepackage{orcidlink}

\DeclareSIUnit\angstrom{\text{\AA}}

\newcounter{subfigure}[figure]

\newcommand{\figurepanelspreamble}{%
  \setcounter{subfigure}{0}%
  \refstepcounter{figure}%
}
\newcommand{\figurepanelspostamble}{\addtocounter{figure}{-1}}
\newcommand{\cappanel}[1]{\textbf{(#1)}}
\newcommand{\alttext}[1]{}

\newcommand{\widepanelref}[4]{%
  \begin{minipage}[t]{#1}
    \vspace{0pt}%
    \refstepcounter{subfigure}%
    \label{#4}%
    \noindent{\sffamily\bfseries\footnotesize(#2)}\par\vspace{0.15em}%
    \includegraphics[width=\linewidth]{#3}%
  \end{minipage}%
}
\newcommand{\widepaneloverview}[4]{%
  \begin{minipage}[t]{#1}
    \vspace{0pt}%
    \refstepcounter{subfigure}%
    \label{#4}%
    \noindent{\sffamily\bfseries\footnotesize(#2)}\par
    \centering\includegraphics[width=0.9\linewidth]{#3}%
  \end{minipage}%
}
\newcommand{\panelref}[4]{%
  \begin{minipage}[t]{#1}
    \vspace{0pt}%
    \refstepcounter{subfigure}%
    #4%
    \noindent{\sffamily\bfseries\footnotesize(#2)}\par\vspace{0.15em}%
    \includegraphics[width=\linewidth,trim=8 8 8 8,clip]{#3}%
  \end{minipage}%
}

\begin{document}

\twocolumn[%
\begin{center}
{\Large\bfseries Matcha: Multi-Stage Riemannian Flow Matching for Accurate and Physically Valid Molecular Docking\par}
\vspace{0.7em}
{\normalsize
Daria Frolova\protect\orcidlink{0000-0001-8368-9015}$^{1,2,\dagger}$ \quad
Talgat Daulbaev\protect\orcidlink{0009-0000-3364-8979}$^{1,2,\dagger}$ \quad
Egor Sevriugov\protect\orcidlink{0000-0001-7107-7732}$^{1,2}$ \quad
Sergei A.\ Nikolenko\protect\orcidlink{0000-0003-1150-9390}$^{1}$\\[0.35em]
Dmitry N.\ Ivankov\protect\orcidlink{0000-0002-8224-4118}$^{1,3}$ \quad
Ivan Oseledets\protect\orcidlink{0000-0003-2071-2163}$^{2,4}$ \quad
Marina A.\ Pak\protect\orcidlink{0000-0003-1075-6509}$^{1}$\par}
\vspace{0.45em}
{\small
$^{1}$Ligand Pro, Moscow, Russia\\
$^{2}$Artificial Intelligence Center, Skolkovo Institute of Science and Technology, Moscow, Russia\\
$^{3}$Center for Bio- and Medical Technologies, Skolkovo Institute of Science and Technology, Moscow, Russia\\
$^{4}$Artificial Intelligence Research Institute, Moscow, Russia\\[0.2em]
$^{\dagger}$These authors contributed equally.
Correspondence: \href{mailto:d.frolova@ligandpro.ru}{d.frolova@ligandpro.ru}.\par}
\end{center}
\vspace{0.55em}
{\small
\noindent\textbf{Abstract.}
Predicting a protein--ligand binding pose is central to structure-based drug design, yet current docking methods still struggle to combine speed, accuracy and physical plausibility.
We introduce \textsc{Matcha}, which refines a ligand pose by three flow-matching stages in translation, rotation and torsion space, then minimizes, filters and ranks candidates with \textsc{GNINA}.
Across the benchmarks considered, \textsc{Matcha} attains the strongest physical plausibility and runs about $28\times$ faster than modern large-scale co-folding models.
Code and benchmarking artifacts are available at \url{https://github.com/LigandPro/Matcha} and archived at \url{https://doi.org/10.5281/zenodo.23568093}.
Model checkpoints are archived at \url{https://doi.org/10.5281/zenodo.23569055}.
Derivations, the training algorithm and additional benchmarks are included as appendices.
\par}
\vspace{0.35em}
{\small\noindent\textbf{Keywords:} molecular docking, flow matching, structure-based drug design, protein--ligand binding pose prediction, deep learning, physical plausibility\par}
\vspace{1.1em}
]

\section{Introduction}
\label{sec:intro}

Drug discovery is the process of identifying molecules that can modulate biological targets involved in disease.
A central difficulty in this process is the enormous chemical search space: even before experimental testing, researchers must decide which candidate molecules are most likely to bind to a target protein and exhibit desirable biological activity~\cite{lipinski2004navigating}.

\begin{figure*}[!tp]
  \centering
  \figurepanelspreamble
  \widepaneloverview{\textwidth}{a}{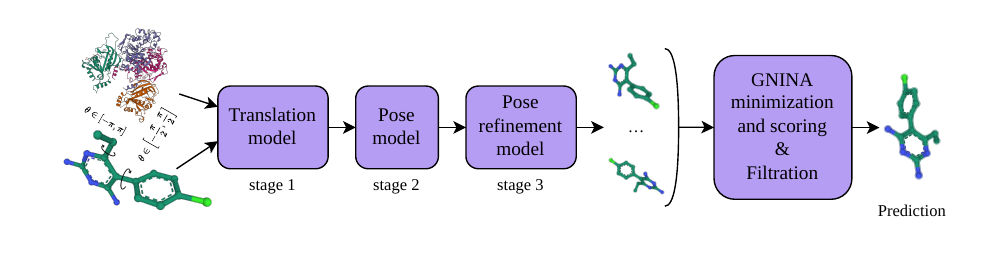}{fig:pipeline}
  \widepaneloverview{\textwidth}{b}{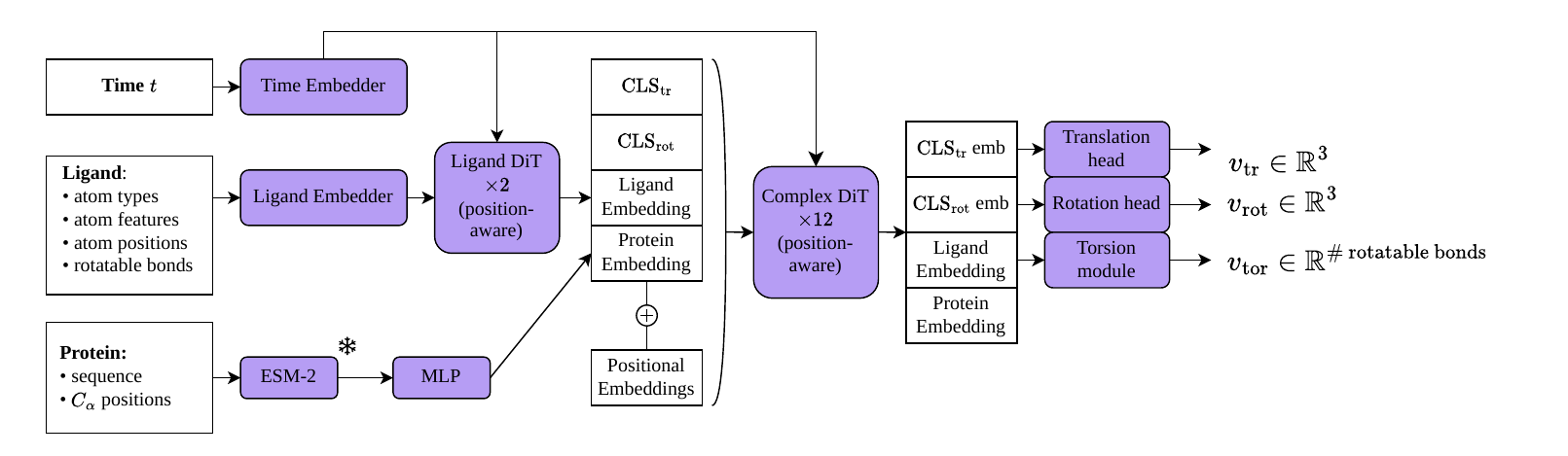}{fig:architecture}
  \figurepanelspostamble
  \caption{
  \textbf{\textsc{Matcha} overview.}
  \cappanel{a} Multi-stage docking pipeline. \textsc{Matcha} uses three flow-matching stages to generate candidate ligand poses, followed by \textsc{GNINA} minimization, unsupervised plausibility filtering and \textsc{GNINA}-based pose selection. Rotatable bonds and their periods define the torsional component of the generative state space.
  \cappanel{b} Architecture of the \textsc{Matcha} velocity model used in stages~1--3. Ligand atoms, protein residues and dedicated translation and rotation tokens are processed by a transformer with distance-aware attention biases, followed by specialized heads for translation, rotation and torsion prediction.
  }
  \label{fig:matcha_overview}
  
\end{figure*}
Computational methods are therefore widely used to prioritize compounds, reduce the number of expensive laboratory experiments, and guide the design of new molecules~\cite{sadybekov2023computational}.

Machine learning has become increasingly important in this setting.
Recent models can predict molecular properties, generate candidate compounds, model protein structures, and reason about protein--ligand interactions~\cite{tropsha2024integrating,jumper2021highly,abramson2024accurate}.
Among these tasks, molecular docking occupies a particularly important role in structure-based drug design.
Given a three-dimensional protein structure and a candidate small molecule, docking aims to predict how the molecule binds inside the protein binding site.
The predicted binding pose can then be used to rank compounds in virtual screening campaigns, inspect molecular interactions, and support downstream medicinal chemistry decisions.

Classical docking approaches~\cite{friesner2004glide,trott2010autodock,koes2013lessons,forli2016computational} have traditionally relied on hand-crafted scoring functions combined with heuristic search algorithms.
However, recent benchmarks~\cite{morehead2026assessing} demonstrate that such methods are increasingly outperformed by data-driven approaches.

Most modern data-driven blind docking methods~\citep{abramson2024accurate,chai2024chai,wohlwend2025boltz}, starting from the seminal \textsc{DiffDock}~\citep{corso2023diffdock}, typically formulate molecular docking as a generative modeling problem.
In this view, a neural network learns to sample plausible ligand poses conditioned on the protein structure and the ligand geometry.

Our proposed method, \textsc{Matcha}, follows this generative paradigm but is based on flow matching~\citep{lipman2023flow} rather than diffusion.
Following \textsc{DiffDock}, we represent ligand flexibility in a joint space of translations, global rotations, and internal torsions.
This corresponds to a semi-flexible ligand: the conformation is fixed except for rotations around rotatable bonds.
In contrast to Riemannian diffusion-based methods, Riemannian flow matching~\citep{chen2024flow} provides tractable losses in these spaces and simplifies training.
Moreover, our approach naturally bypasses the need for semi-flexible conformational alignment, which is a challenging optimization problem.

Architecturally, \textsc{Matcha} consists of three neural networks, trained at different noise levels.
For instance, the first model is optimized to predict the 3D translational displacement of the ligand center relative to the binding site.
By default, \textsc{Matcha} is trained for blind docking, but the pocket-informed setting can be managed by omitting the coarse model and providing the correct binding site location.

\textsc{Matcha} is approximately $28\times$ faster than \textsc{AlphaFold~3}, \textsc{Chai-1} and \textsc{Boltz-2}, while achieving state-of-the-art docking performance. For example, on the \textsc{Astex} test set $89.2\pm1.5$\% of predictions satisfy both $\mathrm{RMSD}\leq\SI{2}{\angstrom}$ and PoseBusters~\citep{buttenschoen2024posebusters} physical-validity criteria (PB-valid), reported as mean\,$\pm$\,s.d. over five inference seeds. 
Across all benchmarks, \textsc{Matcha} also achieves best-in-class physical plausibility.

\section{Results and Discussion}
\label{sec:main_results}
\subsection{Overview of Matcha}

Figure~\ref{fig:matcha_overview} summarizes the end-to-end workflow and model architecture.
\textsc{Matcha} works in ligand translation, global rotation and torsional degrees of freedom. Stage~1 gives a coarse ligand location. Stage~2 tightens the translation while still exploring rotations and torsions. Stage~3 is the low-noise refinement step. 
The candidates then go through the post-processing stage that involves GNINA minimization, a small PoseBusters-inspired validity filter and GNINA-based pose ranking. 
We train no pose scorer of our own: the final selection reuses the pretrained \textsc{GNINA} CNN predictors (see Eq.~\ref{eq:gnina_scoring}).

The velocity model is a Diffusion Transformer (\textsc{DiT})~\cite{peebles2023scalable} with spatial encoder inspired by \textsc{UniMol} and \textsc{AlphaFold~3} \cite{zhou2023uni,abramson2024accurate}. Protein residues are initialized with ESM-2 embeddings \cite{lin2022language}; ligand atoms are initialized from RDKit features \cite{landrum2024rdkit}. 
Pairwise geometry is incorporated through an additional bias term in the attention module. 
Separate heads predict the translational, rotational and torsional velocities used by the Riemannian Euler solver.

\subsection{Evaluation scope}
We compare \textsc{Matcha} and the faster \textsc{Matcha latte} to the baselines using two metrics: symmetry-corrected ligand heavy-atom RMSD and the stricter PoseBusters validity success rate (``PB-valid'') \cite{buttenschoen2024posebusters}.

We report results for four datasets: 
\begin{itemize}
    \item \textsc{Astex}~\cite{hartshorn2007diverse} test set (Figure~\ref{fig:res_astex}, $n=85$),
    \item \textsc{PoseBusters V2}~\cite{buttenschoen2024posebusters} set (Figure~\ref{fig:res_posebusters}, $n=308$),
    \item the time-split \textsc{PDBBind}~\cite{liu2015pdb} test set (Figure~\ref{fig:res_pdbbind}, $n=363$), 
    \item the out-of-distribution \textsc{DockGen}~\cite{corso2024discovery} set (Figure~\ref{fig:res_dockgen}, $n=330$).
\end{itemize}

The baselines include classical rigid docking methods (\textsc{SMINA}~\cite{koes2013lessons}, \textsc{VINA}~\cite{trott2010autodock}, \textsc{GNINA}~\cite{mcnutt2021gnina}), deep learning rigid models (\textsc{DiffDock}~\cite{corso2023diffdock}, \textsc{Uni-Mol}~\cite{zhou2023uni}, \textsc{NeuralPlexer}~\cite{qiao2024state}, \textsc{FlowDock}~\cite{morehead2025flowdock}, \textsc{DynamicBind}~\cite{lu2024dynamicbind}, and \textsc{SurfDock}~\cite{cao2025surfdock}) and large co-folding systems (\textsc{AlphaFold~3}~\cite{abramson2024accurate}, \textsc{Chai-1}~\cite{chai2024chai}, \textsc{Boltz-2}~\cite{passaro2025boltz}). Software versions and inference settings are in Methods~\ref{ssec:benchmark_protocol}.

\begin{figure*}[!tp]
  \centering
  \figurepanelspreamble

  \widepanelref{\textwidth}{a}{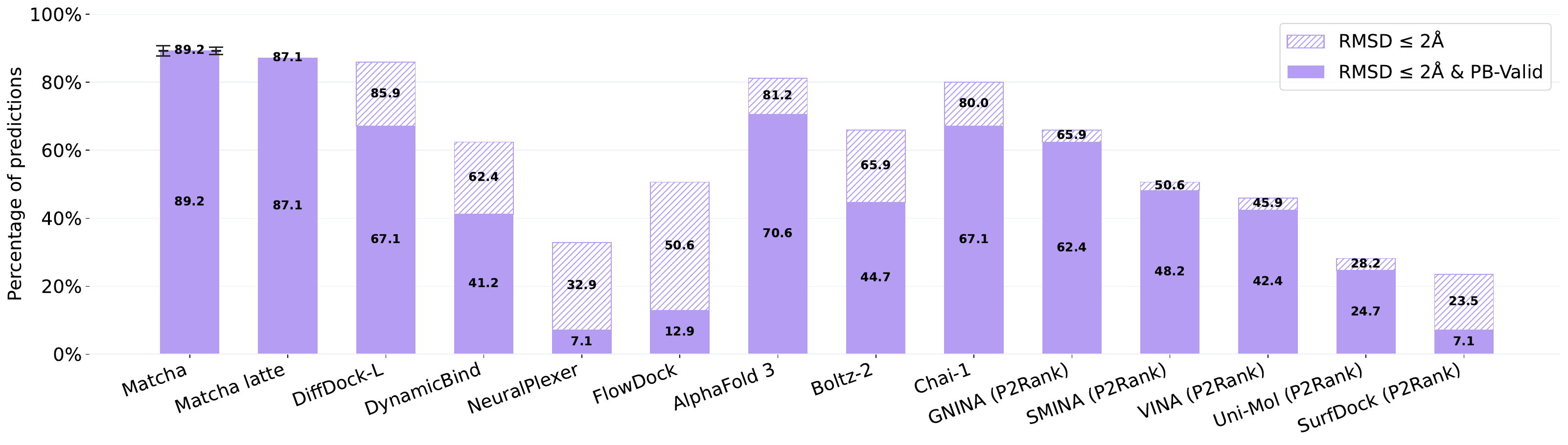}{fig:res_astex}

  \vspace{0.8em}

  \widepanelref{\textwidth}{b}{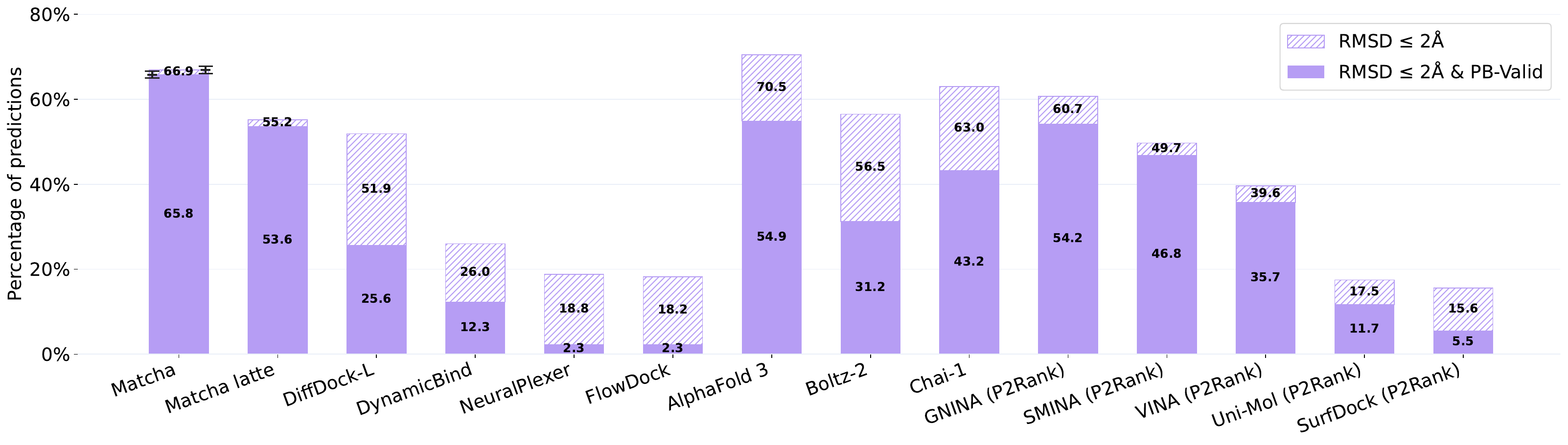}{fig:res_posebusters}

  \vspace{0.8em}

  \widepanelref{\textwidth}{c}{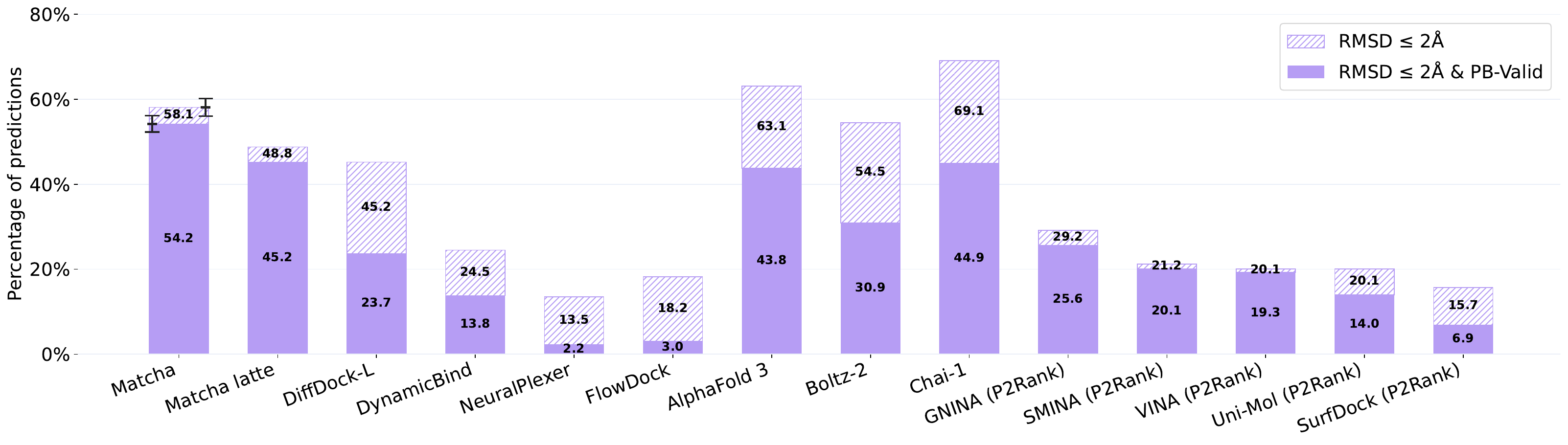}{fig:res_pdbbind}

  \vspace{0.8em}

  \widepanelref{\textwidth}{d}{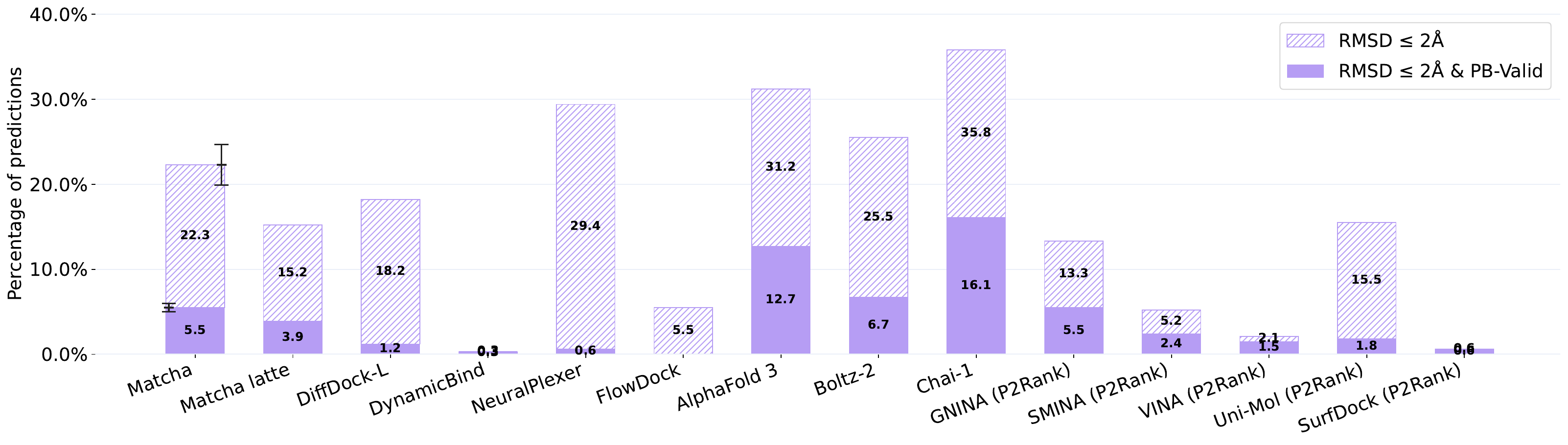}{fig:res_dockgen}

  \figurepanelspostamble
  \caption{
  \textbf{Blind ligand docking performance across benchmarks.}
  \cappanel{a} \textsc{Astex} diverse set ($n=85$).
      \cappanel{b} \textsc{PoseBusters V2} benchmark set ($n=308$).
  \cappanel{c} \textsc{PDBBind} test set ($n=363$).
  \cappanel{d} \textsc{DockGen} dataset ($n=330$).
  For \textsc{Matcha}, bars show the mean over five inference seeds, and the error bars are one standard deviation across those seeds. The interval to the left of each \textsc{Matcha} bar is the variability of the PB-valid rate; the interval to the right is the variability of the $\mathrm{RMSD}\leq\SI{2}{\angstrom}$ rate.
  }
  \label{fig:benchmark_main_results}
  
\end{figure*}

\subsection{Benchmark performance across diverse docking settings}

\subsubsection{Blind setup}

Blind molecular docking aims to predict a ligand binding pose without knowing the binding-site location in advance. The method must therefore identify a suitable pocket on the protein surface and recover the ligand translation, orientation and torsional state within that pocket.
This couples pocket localization with pose generation: a geometrically reasonable ligand conformation can still be counted as incorrect if it is placed in the wrong site, and a low-RMSD prediction can still be unsuitable if it violates basic physical constraints. 
For this reason, we evaluate blind docking using both symmetry-corrected RMSD and the stricter PoseBusters-valid success rate.

On the \textsc{Astex} dataset (Figure~\ref{fig:res_astex}), \textsc{Matcha} demonstrates superior performance in the blind docking setup across all metrics, achieving the highest success rates for both $\mathrm{RMSD}\leq\SI{2}{\angstrom}$ and the harder $\mathrm{RMSD}\leq\SI{2}{\angstrom}$ \& PB-valid ($89.2\pm1.5$\%).
On the physically valid metric, this is 18.6\% above the next-best baseline, \textsc{AlphaFold~3} (70.6\%), and 22.1\% above \textsc{DiffDock-L} (67.1\%). 
It is worth mentioning that all our poses pass PoseBusters filters.
Success rates for \textsc{Matcha} are averaged over five inference seeds, and differences of about 1\% between methods are comparable to this seed-to-seed variability.

On \textsc{PoseBusters V2} (Figure~\ref{fig:res_posebusters}), \textsc{Matcha} achieves $65.8\pm0.8$\% on $\mathrm{RMSD}\leq\SI{2}{\angstrom}$ \& PB-valid and maintains superior success rates over other DL-based docking methods (\textsc{DiffDock-L}, \textsc{NeuralPlexer}, \textsc{FlowDock}) on the full set ($n=308$).
On the temporally held-out subset ($n=130$), co-folding performance drops while \textsc{Matcha} and other DL-based and classical methods remain stable, indicating that our results are robust under fair evaluation; a detailed analysis of train--test overlap is provided in Appendix~\ref{app:additional_results} (Fig.~\ref{fig:posebusters}).

On the \textsc{PDBBind} time-split test set, \textsc{Matcha} again has the highest physically valid success rate among the evaluated methods (Fig.~\ref{fig:res_pdbbind}). Some co-folding systems reach higher raw $\mathrm{RMSD}\leq\SI{2}{\angstrom}$ success, but \textsc{Matcha} has the best $\mathrm{RMSD}\leq\SI{2}{\angstrom}$ \& PB-valid score: $54.2\pm1.9$\%, compared with 44.9\% for the next-best baseline, \textsc{Chai-1}.
This demonstrates that \textsc{Matcha} produces the most physically plausible and chemically valid structures among the evaluated methods on this benchmark.

The \textsc{DockGen} benchmark contains 330 complexes whose binding pockets are structurally dissimilar to pockets seen during \textsc{Matcha} training. 
All methods degrade on this set, including \textsc{Matcha}, which reaches $5.5\pm0.5$\% on $\mathrm{RMSD}\leq\SI{2}{\angstrom}$ \& PB-valid (Figure~\ref{fig:res_dockgen}). Co-folding systems keep an advantage, but still perform poorly.
This comparison is confounded by training-data exposure. \textsc{AlphaFold~3} and the other co-folding models were trained on complexes deposited up to their structure cutoff, not only on proteins; for \textsc{AlphaFold~3} that cutoff is September 2021. \textsc{DockGen} is assembled from \textsc{Binding MOAD} and \textsc{PDBBind} complexes deposited well before that date, so those complexes fall inside the co-folding training window, whereas the corresponding pockets are held out from \textsc{Matcha} training by construction. Higher co-folding success on this set therefore need not reflect better generalization to unseen pockets.
We treat \textsc{DockGen} as an out-of-distribution stress test, not as the same kind of benchmark as \textsc{Astex} or \textsc{PoseBusters}.
From a practical perspective, this benchmark may be of limited relevance, as many of its ligands are not drug-like~\cite{pak2025bento}.

\subsubsection{Pocket-aware setup}

Blind docking evaluates both binding-site localization and pose generation. To isolate the second component, we additionally evaluate all applicable methods in a pocket-aware setting, where the binding-site location is assumed to be known. 
In this protocol, the true binding-site center is provided as input. 
For \textsc{Matcha}, this means that the coarse localization model is omitted and inference starts directly from Stages 2 and 3, initialized at the native pocket center. 
Classical docking tools receive the same center as the search-box origin. 
This setup therefore tests whether a method can recover the correct ligand orientation and torsional state once the binding site has already been identified.

For \textsc{Boltz-2}, we use its pocket-constraint interface and provide residue masks corresponding to direct ligand contacts and extended pocket context, defined by residues within 4 and 10\,\AA\ of the native ligand, respectively. 
We exclude \textsc{Uni-Mol} from this comparison because its standard pocket-aware protocol crops a tight box around the crystal ligand, which provides substantially more spatial information than a center-only initialization and may leak the native pose. 

\begin{figure*}[!t]
  \centering
  \figurepanelspreamble

  \panelref{0.49\textwidth}{a}{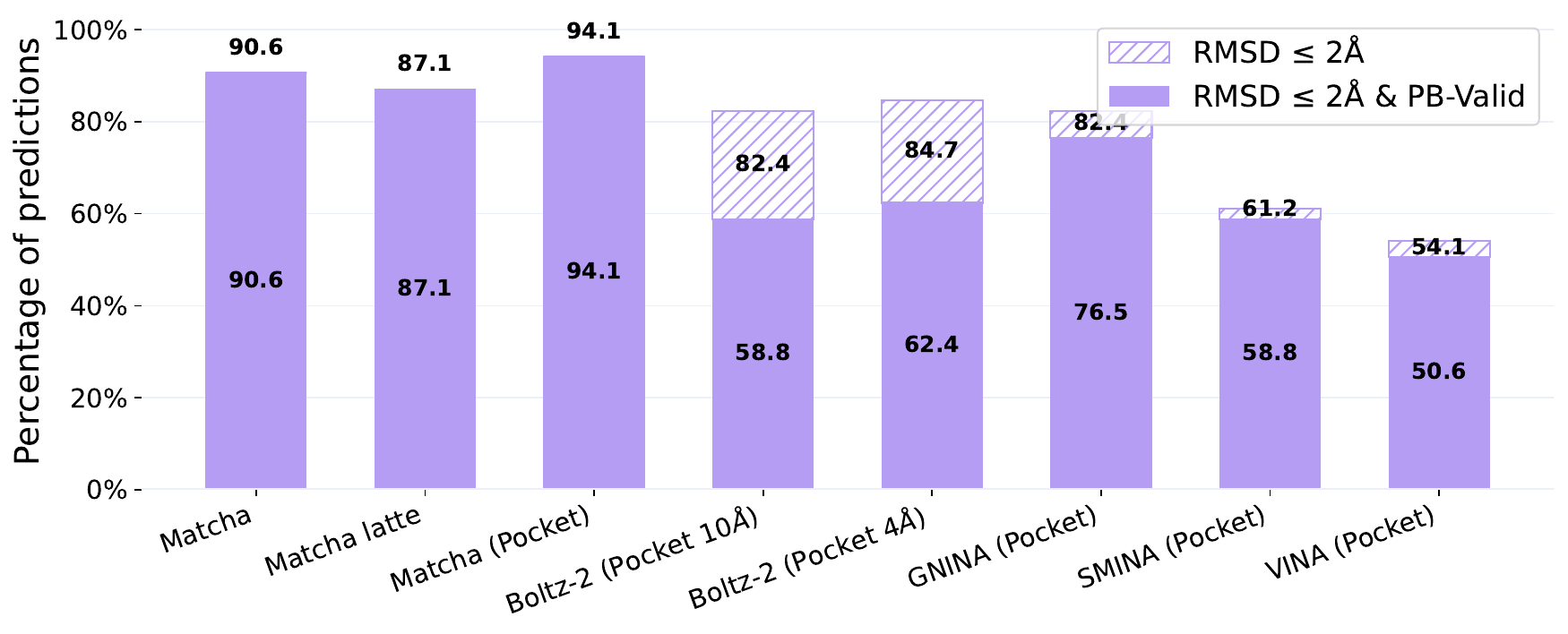}{%
    \label{fig:astex_from_true}%
    \label{fig:pocket_aware_astex_pdbbind}%
  }
  \hfill
  \panelref{0.49\textwidth}{b}{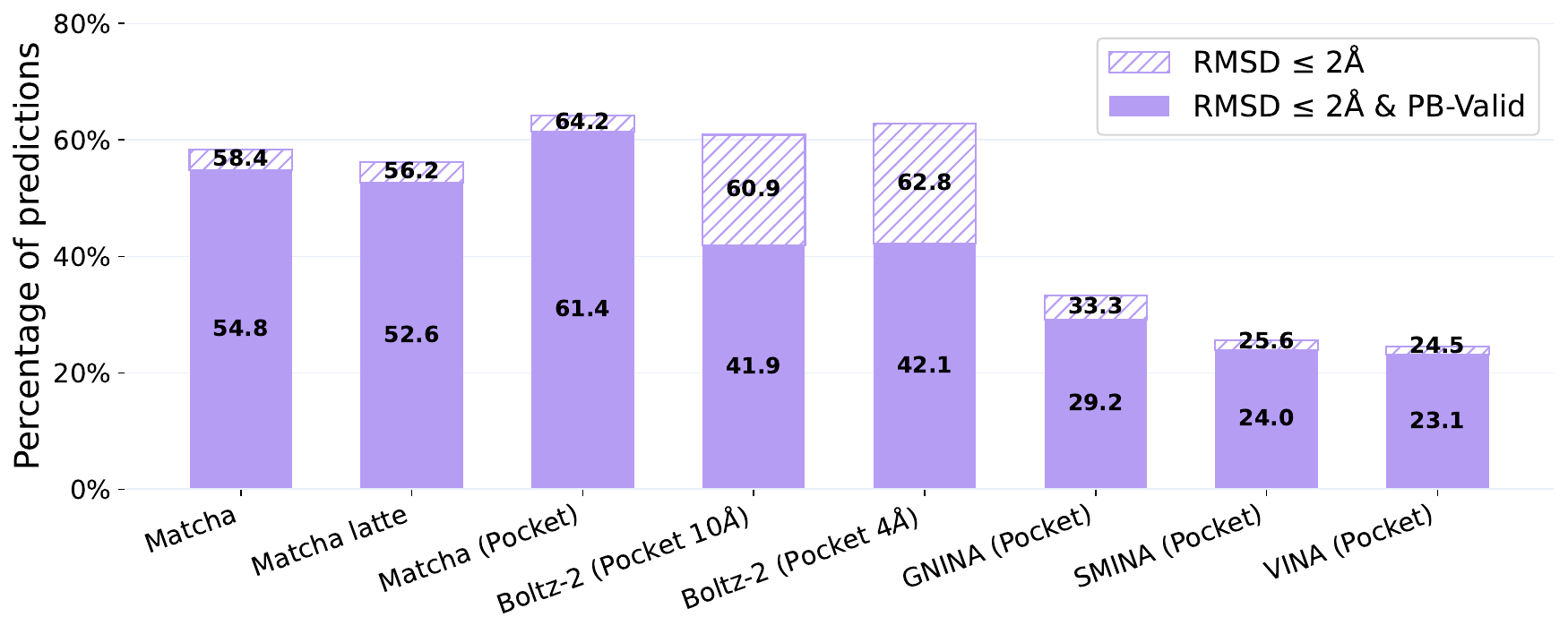}{%
    \label{fig:pdbbind_from_true}%
  }

  \vspace{0.8em}

  \panelref{0.49\textwidth}{c}{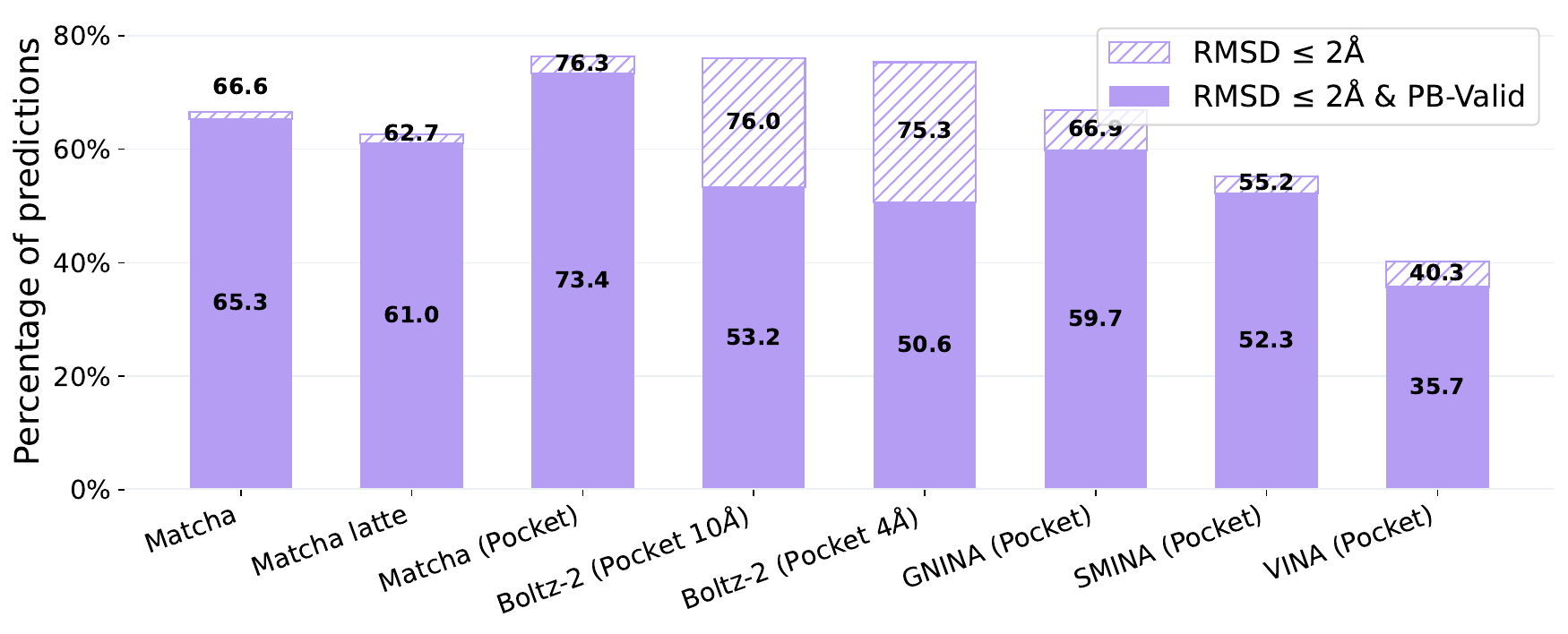}{%
    \label{fig:posebusters_from_true}%
    \label{fig:pocket_aware_posebusters_dockgen}%
  }
  \hfill
  \panelref{0.49\textwidth}{d}{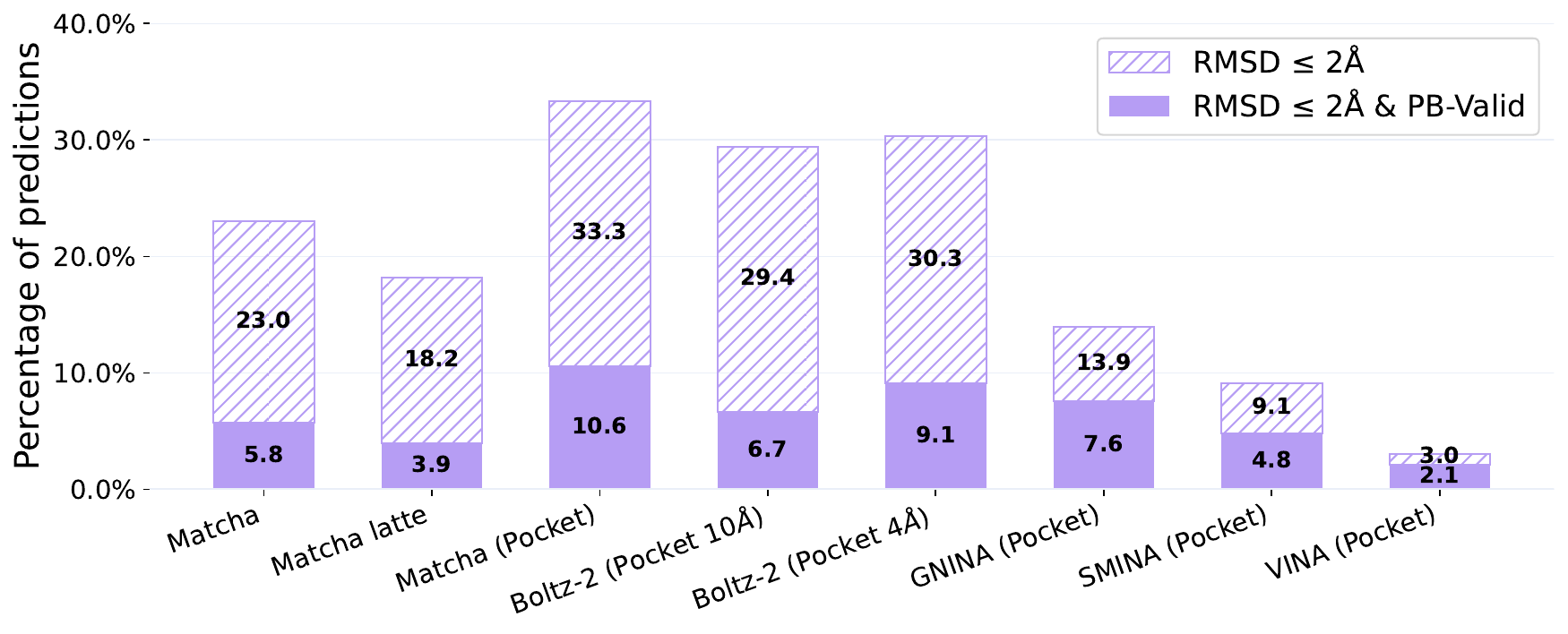}{%
    \label{fig:dockgen_from_true}%
  }

  \figurepanelspostamble
    \caption{
        \textbf{Pocket-aware ligand docking performance.}
        Docking success is evaluated with the true binding-site center provided, separating pose-generation accuracy from pocket-localization accuracy.
        \cappanel{a} \textsc{Astex} diverse set.
        \cappanel{b} \textsc{PDBBind} test set.
        \cappanel{c} \textsc{PoseBusters V2}.
        \cappanel{d} \textsc{DockGen}.
        Success is reported as the percentage of predictions with symmetry-corrected heavy-atom RMSD $\leq\SI{2}{\angstrom}$ and, where indicated, satisfying the PoseBusters validity criteria.
    }
  \label{fig:pocket_aware_all}
  
\end{figure*}

Figure~\ref{fig:pocket_aware_all} shows that \textsc{Matcha} remains competitive in the pocket-aware setting across all four benchmarks. 
Supplying the true pocket center improves the performance of all methods that depend on pocket localization, confirming that binding-site identification is a major source of error in blind docking. 
However, the relative trends remain similar to the blind setup: \textsc{Matcha} achieves strong $\mathrm{RMSD}\leq\SI{2}{\angstrom}$ success rates while maintaining high physical validity. 
In particular, under the stricter $\mathrm{RMSD}\leq\SI{2}{\angstrom}$~\&~PB-valid metric, \textsc{Matcha} consistently outperforms the classical docking baselines. 
This indicates that the learned flow-matching refinement stages contribute beyond pocket localization alone: even when the binding site is known, the model provides an effective proposal distribution over rotations and torsions before all-atom minimization and filtering.

\subsection{Computational efficiency of \textsc{Matcha}}

\textsc{Matcha} is fast enough for screening-scale use.
On a single NVIDIA A100 40GB GPU, the full model runs in about 13.6\,s per complex. 
Timings exclude model loading and include all sampling used for the reported multi-pose selection.

\begin{figure*}[!t]
  \centering
  \includegraphics[width=0.62\textwidth]{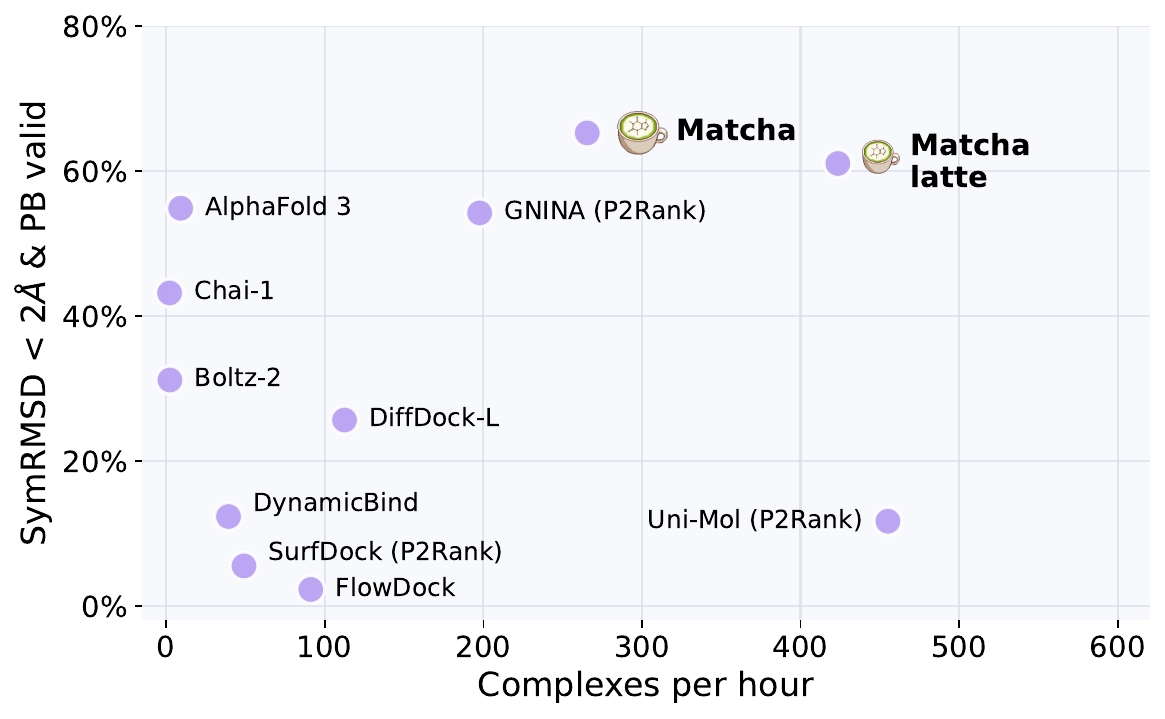}
  \caption{\textbf{
Average inference time on \textsc{Astex} and $\mathrm{RMSD}\leq\SI{2}{\angstrom}$~\&~PB-valid success on \textsc{PoseBusters V2}.}}
  \label{fig:timing}
  
\end{figure*}

Inference time in Figure~\ref{fig:timing} is measured on \textsc{Astex}: runtime is a per-complex cost, and this small diverse set is enough to estimate it.
The success rates in the same figure are on \textsc{PoseBusters V2}, the larger held-out set used for that accuracy comparison.
The exact timing results (in seconds) are detailed in Appendix~\ref{app:timing}.
\textsc{Matcha} demonstrates the best speed-accuracy balance in terms of PB-valid predictions with $\mathrm{RMSD}\leq\SI{2}{\angstrom}$ (see Figure~\ref{fig:timing}), which is important in practical applications.
\textsc{Matcha} achieves substantially higher docking success rates compared to fast methods (\textsc{DiffDock-L}, \textsc{NeuralPlexer}, \textsc{FlowDock}), while also requiring less inference time.
For settings where throughput is critical, we provide \textsc{Matcha latte}: a lighter variant (5 Euler steps, 10 samples) that runs at $\sim$8.5\,s per complex, $1.6\times$ faster than the full \textsc{Matcha} ($\sim$13.6\,s), with a modest accuracy trade-off.
\textsc{Matcha} also shows a significant speed advantage over high-accuracy co-folding methods, achieving comparable docking quality.

\textsc{Matcha} required significantly less computational resources for training compared to competing methods: 35 GPU-days on a single H100 versus 120 GPU-days for \textsc{DiffDock-L} (4× RTX A6000) and 3,840 GPU-days for \textsc{Chai-1} (128× A100).
This highlights \textsc{Matcha}'s efficiency in both training and inference compared to DL-based models and co-folding models.

\subsection{Analysis of \textsc{Matcha} design choices}

\subsubsection{Contribution of individual \textsc{Matcha} components}
\label{ssec:pipeline_ablation}

\paragraph{Choice of Stages}

\begin{figure*}[!tp]
  \centering
  \figurepanelspreamble

  \panelref{0.32\textwidth}{a}{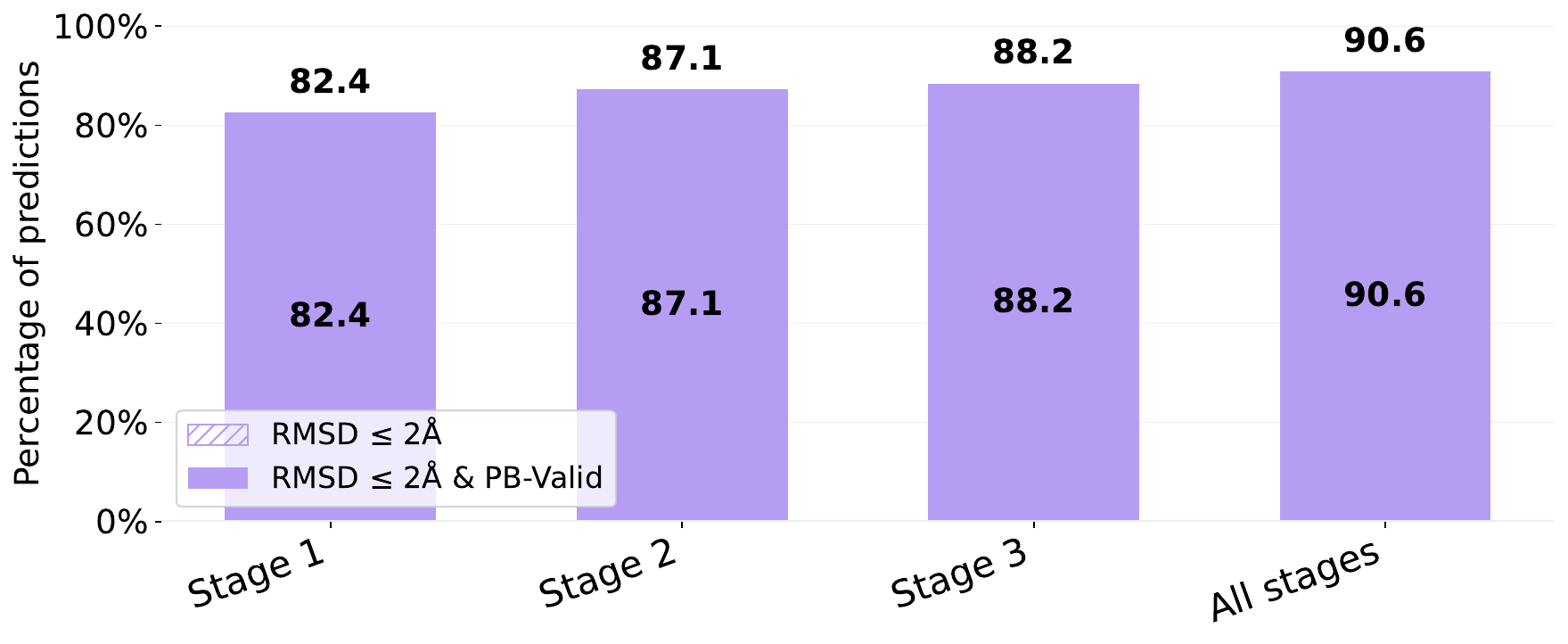}{
    \label{fig:res_pipeline_ablation_astex}
    \label{fig:pipeline_ablation}
    \label{fig:pipeline_and_postprocessing}
  }
  \hfill
  \panelref{0.32\textwidth}{b}{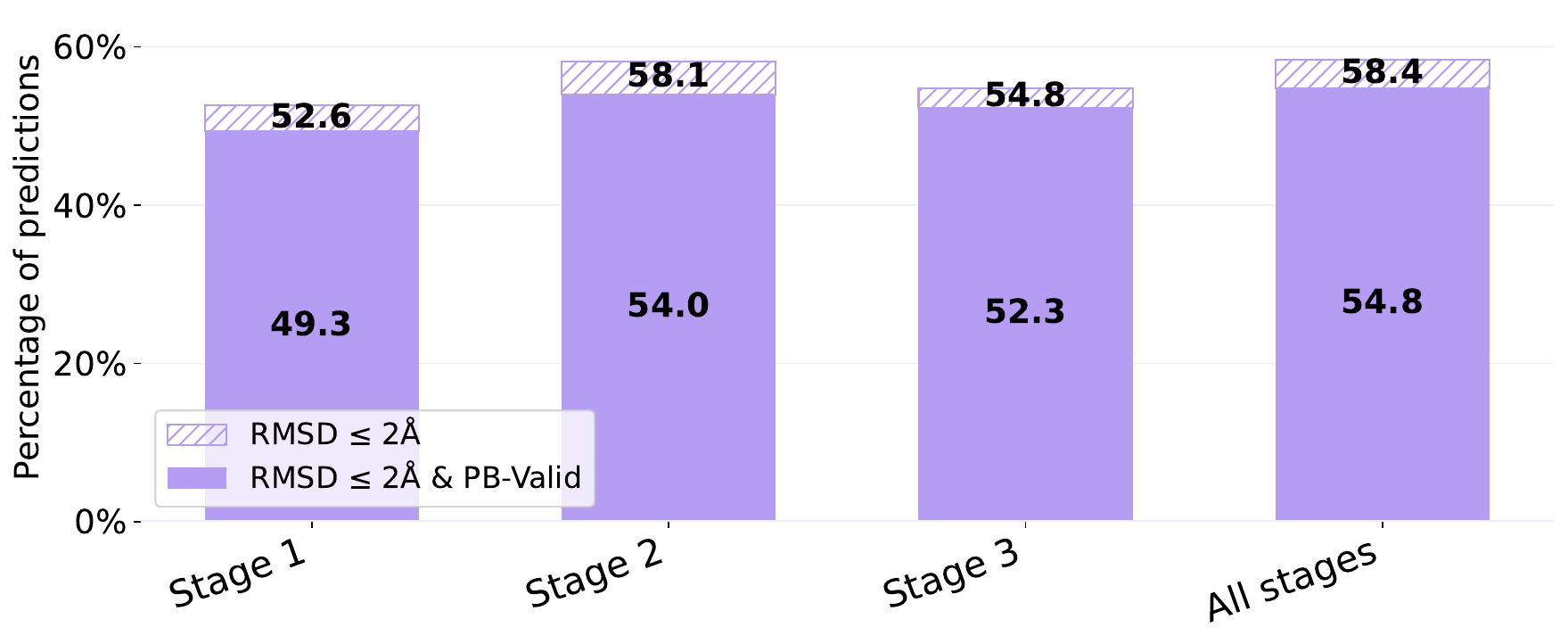}{
    \label{fig:res_pipeline_ablation_pdbbind}
  }
  \hfill
  \panelref{0.32\textwidth}{c}{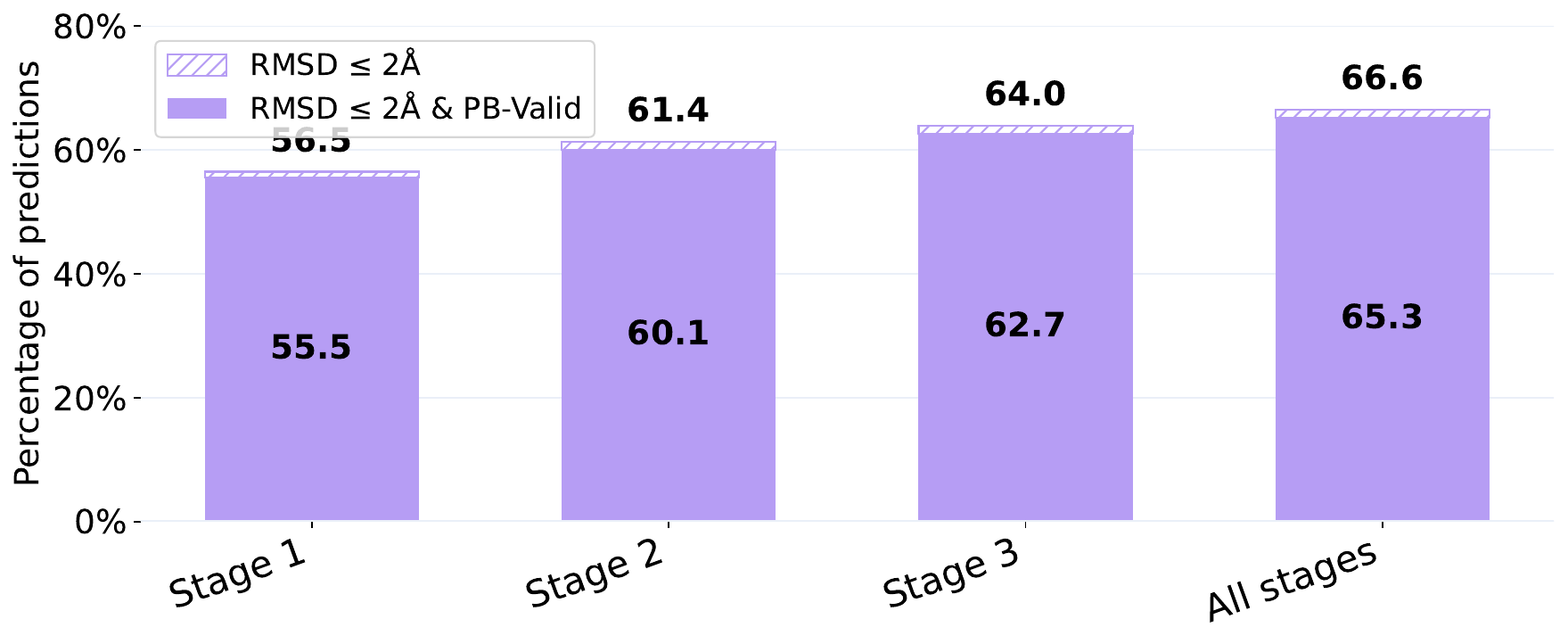}{
    \label{fig:res_pipeline_ablation_posebusters}
  }

  \vspace{0.8em}

  \panelref{0.32\textwidth}{d}{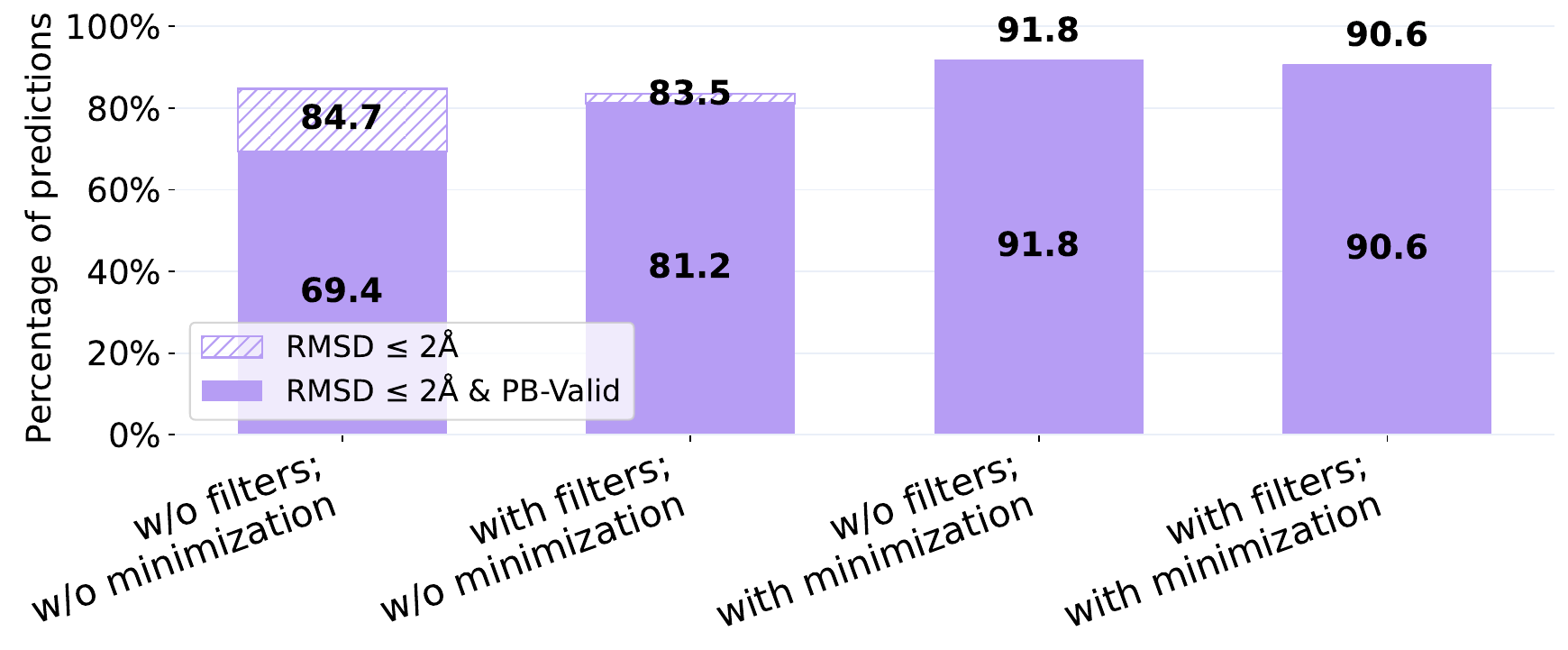}{
    \label{fig:astex_postprocessing}
    \label{fig:astex_pdb_postprocessing}
  }
  \hfill
  \panelref{0.32\textwidth}{e}{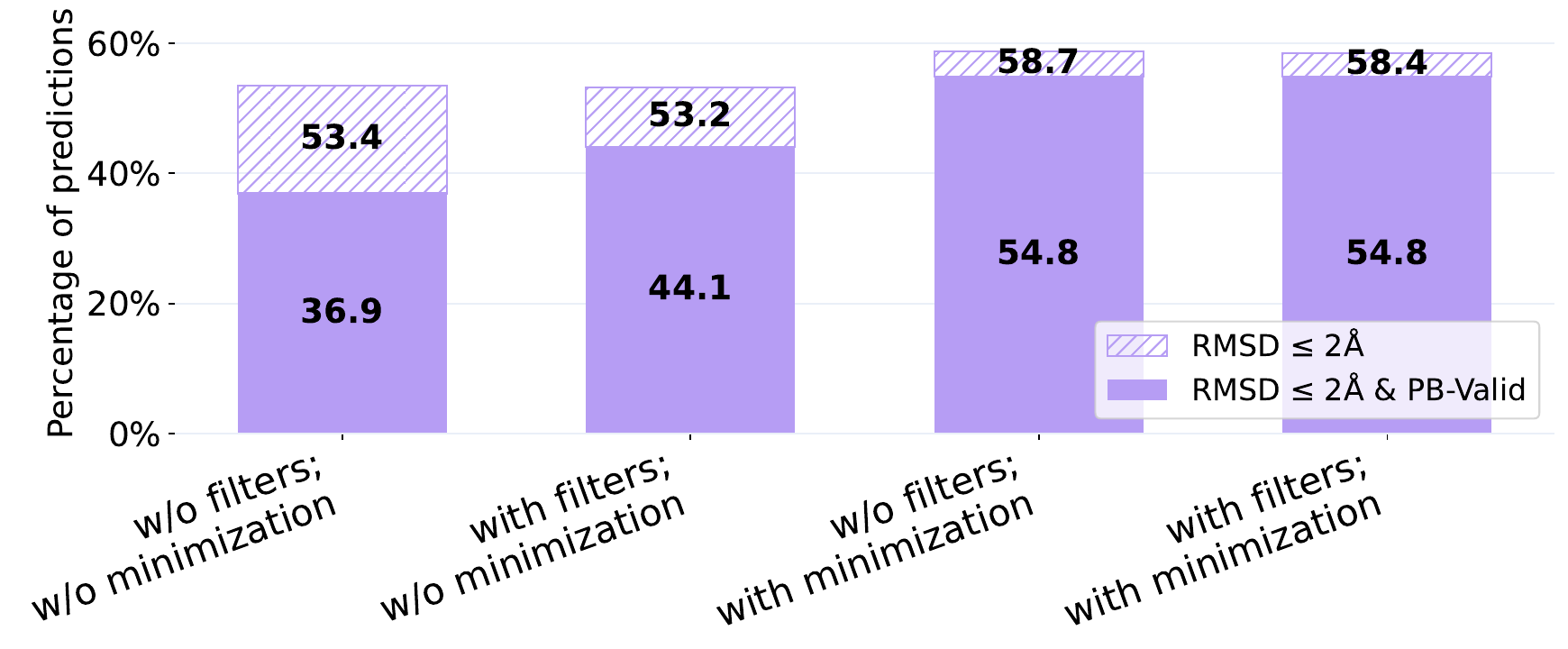}{
    \label{fig:pdbbind_postprocessing}
  }
  \hfill
  \panelref{0.32\textwidth}{f}{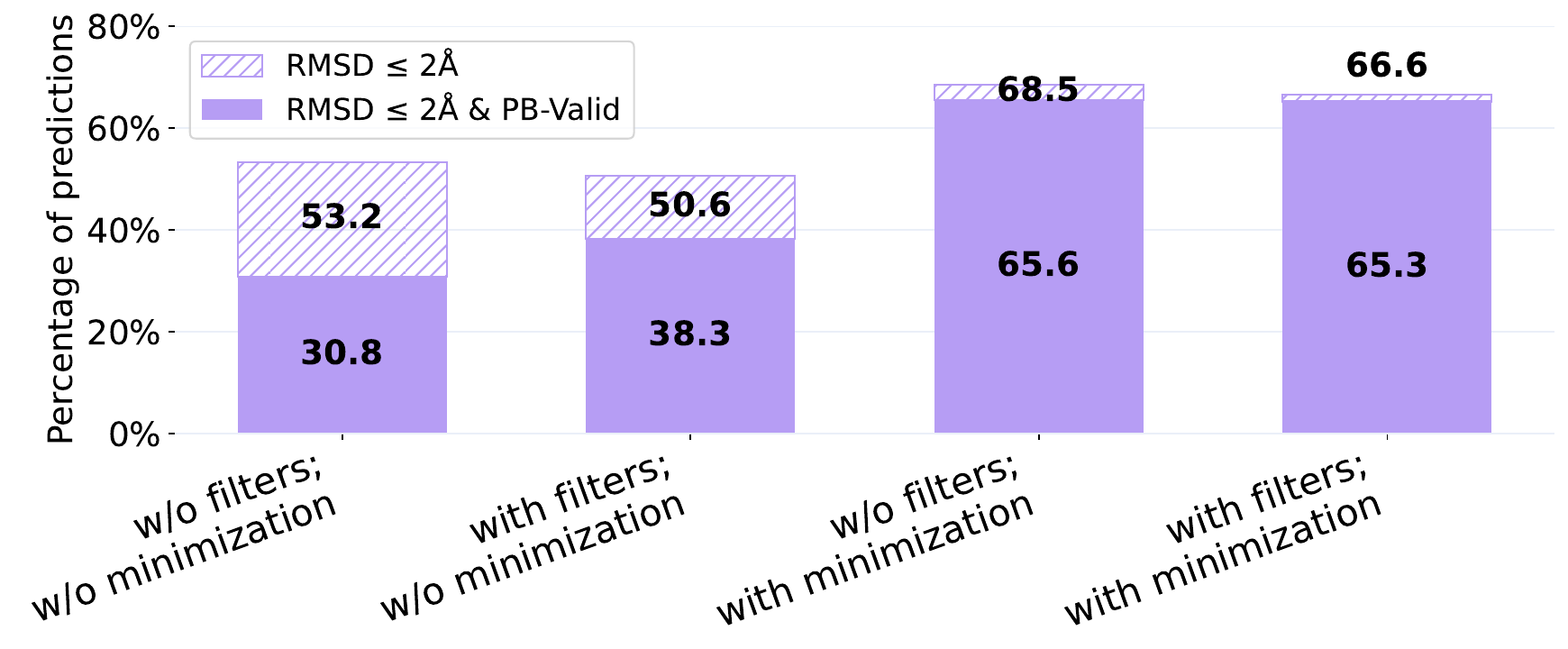}{
    \label{fig:posebusters_postprocessing}
    \label{fig:pb_dg_postprocessing}
    \label{fig:postprocessing}
  }

  \vspace{0.8em}

  \panelref{0.44\textwidth}{g}{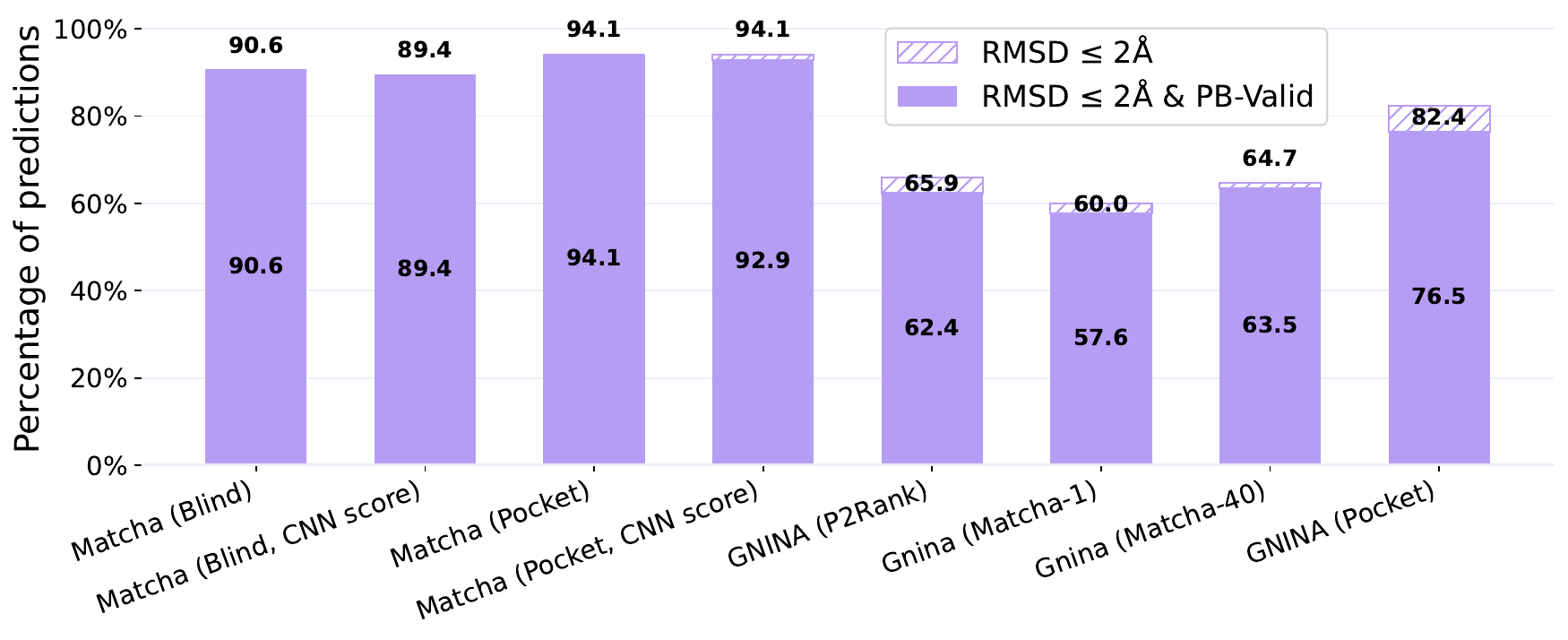}{
    \label{fig:gnina_ablation_astex}
    \label{fig:gnina_and_samples}
  }
  \hspace{0.055\textwidth}
  \panelref{0.44\textwidth}{h}{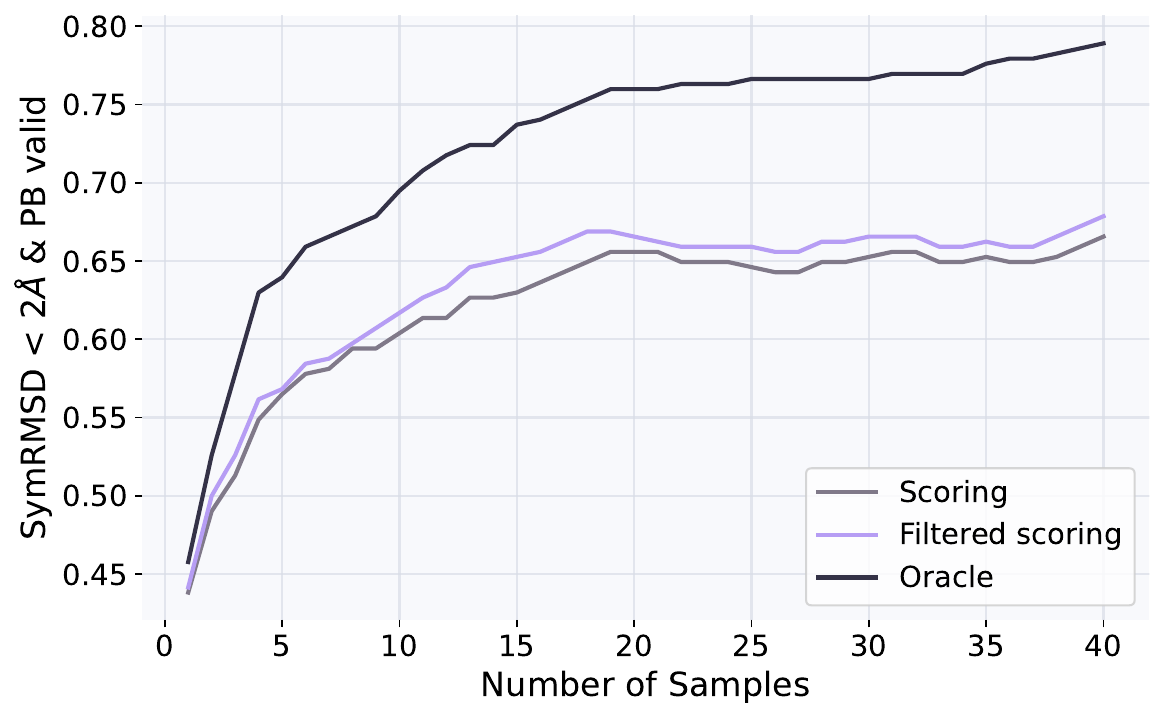}{
    \label{fig:sample_dependence}
  }

  \figurepanelspostamble
  \caption{
  \textbf{Ablation and sampling analyses of \textsc{Matcha}.}
  Blind docking performance is evaluated after progressively varying generation stages, post-processing steps and the number of sampled poses.
  \cappanel{a--c} Stage-wise performance on \textsc{Astex}, \textsc{PDBBind} and \textsc{PoseBusters V2}, respectively.
  \cappanel{d--f} Contribution of post-processing steps on the same benchmarks, respectively.
  \cappanel{g} On \textsc{Astex}, full \textsc{Matcha} (blind and pocket-aware), the same poses ranked by the \textsc{GNINA} CNN score without validity filtering, and \textsc{GNINA} docking from Stage~1 or from a pocket center.
  \cappanel{h} Sample-size dependence on \textsc{PoseBusters V2}.
  Success is reported as the percentage of predictions with symmetry-corrected heavy-atom RMSD $\leq\SI{2}{\angstrom}$ and, where indicated, satisfying the PoseBusters validity criteria.
  }
  \label{fig:ablation_and_additional_analysis}
  \label{fig:matcha_ablation_main}
  
\end{figure*}

We report the results produced by each \textsc{Matcha} stage to demonstrate the effectiveness of a three-stage pipeline design and the improvements produced by each pipeline stage.
The results are reported for \textsc{Astex} and \textsc{PoseBusters V2} datasets in Figure~\ref{fig:pipeline_ablation}.
They demonstrate the effectiveness of a three-stage \textsc{Matcha} design: most complexes are successfully docked within the first two stages.
Stage 3 is an additional low-noise refinement step, while merging predictions from all three stages gives even higher results.

\paragraph{Learned refinement versus GNINA docking}

A natural question is whether the later \textsc{Matcha} stages are necessary, or whether a classical docking engine can recover similar performance given a pocket center from Stage 1. 
To test this, we replace Stages 2 and 3 with \textsc{GNINA} docking. 
Starting from Stage 1 predictions, we initialize GNINA at the predicted ligand center and keep the top-ranked pose according to \textsc{GNINA}'s default CNN score (this baseline replaces Stages~2--3 and does not use the composite scoring rule in Eq.~\ref{eq:gnina_scoring}). 
We evaluate two variants of initialization: from the top Stage 1 translation, denoted \textsc{GNINA (Matcha-1)}, and from all 40 Stage 1 translations, denoted \textsc{GNINA (Matcha-40)}.

Figure~\ref{fig:gnina_ablation_astex} shows that the full \textsc{Matcha} pipeline with three generator stages substantially outperforms these \textsc{GNINA} alternatives, and also default \textsc{GNINA} run from \textsc{P2Rank} or from the true pocket, on both $\mathrm{RMSD}\leq\SI{2}{\angstrom}$ and $\mathrm{RMSD}\leq\SI{2}{\angstrom}$~\&~PB-valid success. 
The same panel includes two bars that keep the \textsc{Matcha} poses but rank them by the default \textsc{GNINA} CNN score alone, without the validity filter: \textsc{Matcha (Blind, CNN score)} and \textsc{Matcha (Pocket, CNN score)}. On this \textsc{Astex} panel their PB-valid rates are 89.4\% and 92.9\%, against 90.6\% and 94.1\% for the default pipeline.
Thus, the gains of \textsc{Matcha} cannot be explained by Stage~1 localization followed by classical docking, nor by the composite score and the validity filter. 
The learned Stage 2 and Stage 3 flow-matching refinements make a major contribution to the final docking quality.

\paragraph{Physical validity assessment}
A serious advantage of \textsc{Matcha} lies in its ability to generate physically plausible molecular poses.
For instance, on the \textsc{PoseBusters V2} test set, \textsc{Matcha} has 98\% of physically-valid structures among those with $\mathrm{RMSD}\leq\SI{2}{\angstrom}$.
To compare, \textsc{AlphaFold~3} has 78\%, \textsc{DiffDock-L} -- 49\%.
This consistent superiority in generating chemically valid poses across different datasets demonstrates that \textsc{Matcha}'s flow matching approach effectively preserves molecular constraints through three key mechanisms: (1) ligand parametrization using only torsional angles, which maintains internal molecular geometry, (2) \textsc{GNINA} minimization, and
(3) validity filtering to eliminate unrealistic complexes.
The high fraction of physically plausible structures, combined with fast inference, is a useful property for screening-scale applications, where chemical validity and throughput matter alongside geometric accuracy.

\paragraph{Contribution of post-processing}

To clarify the contribution of the post-processing pipeline, we ablate it and measure the impact of GNINA minimization and fast filtration (Figures~\ref{fig:astex_pdb_postprocessing} and \ref{fig:pb_dg_postprocessing}).
Even without minimization and filtration, the vanilla \textsc{Matcha} model already achieves a high fraction of physically valid poses, supporting torsion-angle parametrization as a strong design choice for preserving physical plausibility. Adding filtration and GNINA minimization further increases the overall validity.

\subsubsection{Hyperparameters}

\paragraph{Number of samples}

Figure~\ref{fig:sample_dependence} shows how the number of generated samples affects the quality of the best selected pose for the \textsc{PoseBusters V2} dataset.
We consider the minimization-only regime, minimization with filtration (default), and oracle performance (representing the theoretical upper bound).
Using \textsc{GNINA} minimization and validity filtering substantially improves $\mathrm{RMSD}\leq\SI{2}{\angstrom}$ \& PB-valid success rates compared to sampling only one pose. These results demonstrate the importance of incorporating energy minimization.
Performance plateaus at around 20 samples, marking the optimal computational cost-quality trade-off.
The gap between the default scoring pipeline and oracle performance indicates the remaining opportunities for improvement in pose selection.

\paragraph{Euler Solver ablations}

Increasing the number of Riemannian Euler integration steps from 1 to 10 improves docking success rates, after which the gains saturate; we therefore use 10 Euler steps for all main experiments.
The full ablation on \textsc{Astex}, \textsc{PDBBind} and \textsc{PoseBusters V2} is reported in Appendix~\ref{app:additional_results} (Fig.~\ref{fig:euler_steps_extended}).

\subsection{Impact of Alignment Strategy on Reported Metrics}
\label{sec:alignment}

Pocket alignment refers to the procedure for superimposing predicted and reference protein–ligand complexes prior to RMSD evaluation, which is necessary for models that modify protein structure during prediction.
Different strategies lead to systematically different results.
One common option is the \textsc{pocket-based} alignment, where the predicted pocket is aligned to the reference one.
This can yield low RMSD values even if the ligand is placed in a non-native binding site, effectively inflating success rates for methods that predict the protein structure.
In contrast, another strategy is the \textsc{full} alignment, in which the reference pocket is aligned to the full predicted protein, ensuring that incorrect pocket assignments are penalized.
This methodological difference explains why our reported metrics may differ from those in other studies, even when using the same datasets.
We provide details on both alignment strategies as well as their comparison in Appendix~\ref{app:alignment}.
We report results with \textsc{pocket-based} alignment in the main results.

\subsection{Summary of findings}
Overall, \textsc{Matcha} is a multi-stage Riemannian flow matching method for molecular docking that combines geometric pose generation, \textsc{GNINA} minimization, physical validity filtering, and GNINA-based scoring. 
Across diverse benchmarks, it delivers strong docking accuracy together with the highest physical plausibility among the evaluated methods. 
Our results further show that these gains arise primarily from the learned multi-stage refinement, with post-processing providing an additional improvement in robustness. 
Post-processing is nevertheless a substantial component of the pipeline, and applying it to the baseline methods improves them as well; a like-for-like comparison under identical post-processing is reported in Appendix~\ref{app:postprocessing_baselines}, where \textsc{Matcha} remains the strongest method on \textsc{Astex} and is competitive elsewhere at a small fraction of the co-folding inference cost. Fig.~\ref{fig:timing_with_postproc} plots the same post-processed success rates against updated throughput (complexes per hour), which adds the shared post-processing time to each method's inference time.
\textsc{Matcha} is also substantially faster and more computationally efficient than large co-folding systems.
Together, these properties make \textsc{Matcha} attractive for large-scale structure-based workflows, with two caveats that we consider important.
First, like the docking baselines evaluated here, \textsc{Matcha} is benchmarked on holo receptor structures; performance on apo receptors or receptors solved with a different ligand is not measured in this work, and claims about virtual screening should be read with that in mind.
Second, \textsc{DockGen} shows that accuracy drops on pockets that are structurally dissimilar to the training data. Broader receptor coverage and explicit receptor flexibility are the natural next steps.

\section{Methods}

\subsection{Docking loss function}

\textsc{Matcha} treats the receptor as rigid while parameterizing the ligand motion through three components: a three-dimensional translation, a global rotation in $\mathrm{SO}(3)$ and torsional angles in $\mathrm{SO}(2)$ for each rotatable bond. For rotatable bonds, the torsional period $p$ is defined as the smallest positive angle that yields an indistinguishable ligand configuration up to symmetry, so torsion angles are represented modulo $p$ in the interval $\left(-p/2,\,p/2\right]$.

Our model predicts velocities $\vv_\mathrm{tr}$, $\vv_\mathrm{rot}$, and $\vv_\mathrm{tor}$ in the tangent spaces $\mathbb{R}^3$, $\mathfrak{so}(3)$, and $\mathfrak{so}(2)^m$, with $m$ denoting the number of rotatable bonds.
Elements of $\mathfrak{so}(n)$ can be thought of as $n \times n$ real skew-symmetric matrices~\citep{warner1983foundations}, we represent them as $n (n - 1) / 2$-dimensional vectors.

Following conditional flow matching \cite{chen2024flow}, we minimize
\begin{equation}
\mathcal{L}_\mathrm{CFM}=\mathbb{E}_{\vx_0\sim p_0,\vx_1\sim p_\mathrm{data},t\sim \mathrm{U}[0,1]}
\left\lVert \vv_\vtheta(\vx_t,t)-\dot{\vx_t}\right\rVert_g^2,
\end{equation}
where $\vx_t$ is an interpolation between a noisy sample $\vx_0$ and the target pose $\vx_1$. For translations we use linear interpolation with Gaussian noise. For rotations and torsions we use spherical linear interpolation. Torsional motion on $\mathrm{SO}(2)$ is represented by a scalar angular velocity, whereas global rotation on $\mathrm{SO}(3)$ is represented by three coefficients of a skew-symmetric matrix in $\mathfrak{so}(3)$. Rotational and torsional targets are obtained by differentiating the interpolation path, and the corresponding loss terms use the canonical metric $g(\mathbf{X},\mathbf{Y})=\mathrm{tr}(\mathbf{X}^{\top}\mathbf{Y})$ on the tangent space. In practice, this reduces manifold-valued regression targets to a scalar prediction per torsion and a three-parameter prediction for global rotation.

We compute separate flow matching losses~\citep{chen2024flow} for each component and optimize their weighted sum.
Generally, to train a flow matching model $\vv_\vtheta$ given data points $\vx_1$ on a Riemannian manifold $\mathcal{M}$ drawn from $p_\mathrm{data}$, we should define a noise distribution $p_0$.
Then we choose an interpolation on $\mathcal{M}$, select an appropriate norm, and compute the time derivative of the interpolation to obtain the conditional velocity.
The conditional flow matching loss takes the form:
\begin{equation}
    \mathcal{L}_\mathrm{CFM} = \mathbb{E}_{\vx_0 \sim p_0, \vx_1 \sim p_\mathrm{data}, t \sim \mathrm{U}[0, 1]} \|\vv_\vtheta(\vx_t, t) - \dot{\vx_t} \|_g^2,
\end{equation}
where $\vx_t = \mathrm{interpolate}(\vx_0, \vx_1; t)$.
Here $\|\cdot\|_g$ is the norm induced by the Riemannian metric $g$ on the corresponding tangent space, so each component loss reduces to a squared error in tangent-space coordinates (Appendix~\ref{sec:app_so_n}).

For translations, we use a standard linear interpolation and normally distributed noise.
For both angular components, we adopt spherical linear interpolation (SLERP) \cite{shoemake1985animating}.
In $\mathrm{SO}(3)$, the time derivative of SLERP is computed by transforming to quaternion representations and applying automatic differentiation with custom backward functions.
Additional derivation details are provided in Appendix~\ref{sec:app_so_n}.

\subsection{Neural network architecture}

\textsc{Matcha} consists of two primary components (Figure~\ref{fig:pipeline}): a docking module and a physically-aware post-processing module that performs \textsc{GNINA} energy minimization, physical validity filtering, and GNINA-based pose ranking.
The velocity model architecture is shown in Figure~\ref{fig:architecture}.

\subsubsection{Velocity model}
\label{ssec:velocity_model}
\paragraph{Input tokens}
The input sequence consists of ligand atom tokens, protein amino acid residue tokens, and two CLS-like tokens for aggregating translation ($\mathrm{CLS}_\mathrm{tr}$) and global rotation ($\mathrm{CLS}_\mathrm{rot}$) information.
Each token is assigned a 3D coordinate: atom positions for ligand atoms, $C_\alpha$ positions for residues, and the ligand centroid for both CLS tokens.
Protein representations are initialized from ESM-2-35M~\citep{lin2022language} embeddings; we use the 35M model instead of the 650M variant to reduce the risk of overfitting.
Initial ligand atom embeddings are a sum of simple embeddings of scalar and categorical atom features from the RDKit package~\citep{landrum2024rdkit}.

Time $t\!\in\![0,1]$ is embedded by an MLP over sinusoidal features and conditions all transformer blocks in a \textsc{DiT}-like manner.
Positions $(x, y, z)$ are encoded using a simple MLP and added to both embeddings of ligand atoms and protein residues in a manner of positional encoding in transformers.

\paragraph{Distance-aware attention bias}
We adopt the approach from \textsc{UniMol} and \textsc{AlphaFold~3}, where pairwise spatial features enter self-attention as additive per-head biases. For each token pair we combine a radial term (a Gaussian-RBF embedding of a stabilized, edge-type--specific inverse distance) with a directional term (a learned projection of the displacement vector); the resulting per-head bias is added to the attention logits. The full parameterization is given in Appendix~\ref{app:attention_bias}.

\paragraph{Velocity prediction heads}
After the transformer backbone, we employ lightweight modules to predict the velocity fields.
We do not force rotational or translational symmetries via the architecture; instead, we rely on data augmentations during training to promote invariances and equivariances.

The translation head consumes the dedicated $\mathrm{CLS}_{\mathrm{tr}}$ token and outputs a 3D velocity vector $v_{\mathrm{tr}} \in \mathbb{R}^3$.
The rotation head consumes the $\mathrm{CLS}_{\mathrm{rot}}$ token and outputs a 3-vector representation of $\mathfrak{so}(3)$.
For torsions, we construct a token for each rotatable bond by averaging the embeddings of ligand atoms influenced by the rotation of that bond and combining this with an encoding of basic bond-level features.
The resulting per-bond sequence is passed through a lightweight transformer decoder, and a final single-layer MLP projects each token to a scalar torsional velocity $v_{\mathrm{tor}}$.

\paragraph{Coarse-to-fine structure}
\textsc{Matcha} consists of three models with identical architecture but independently trained weights.
The first model is used solely for translation, where samples are drawn from a zero-mean Gaussian distribution with a large variance, while angular components are sampled uniformly.
The second model refines translation using a Gaussian centered at the ground truth with moderate variance, still keeping angular components uniform.
Finally, the third model sharpens both translation and angular degrees of freedom, sampling them from Gaussians with small variance around the ground truth.
The models are trained independently.
Full details are provided in Appendix~\ref{app:training_algorithm} (Algorithm~\ref{alg:training}).

The system can operate in two distinct scenarios: blind docking and pocket-aware docking.
Blind docking means predicting ligand poses without prior knowledge of the binding site location, while in the pocket-aware scenario, the information about the known binding site is used to guide pose prediction.
This flexibility is achieved because of the multiscale nature of \textsc{Matcha}.

\paragraph{Augmentations}
\label{sec:aug}
During training, we apply multiple augmentation techniques to avoid overfitting and improve model generalization.
Firstly, we randomly rotate the whole complex to get new $(x, y, z)$ coordinates.
Secondly, we add random Gaussian noise with zero mean and standard deviation 0.25 to the protein and ligand positions. We also add zero-mean Gaussian noise with std 0.1 to the ESM embeddings of protein residues.
Finally, we randomly mask $15\%$ of protein residues and ligand atoms.
This strategy also leads to masking of some rotatable bonds.

\paragraph{Inference}
We run a fixed-length explicit Riemannian Euler solver (10 steps) over $(\mathrm{tr},\mathrm{rot},\mathrm{tor})$ using the predicted velocities, applying three models sequentially.

\begin{itemize}
\item \textbf{Stage 1.} From a random initialization, we integrate all degrees of freedom (translation, rotation, torsions), but retain only the predicted translation; the angular components are discarded and reinitialized uniformly.
\item \textbf{Stage 2.} Starting from this state (predicted translation and uniformly distributed angles), we perform the same rollout and pass the full output $(\mathrm{tr},\mathrm{rot},\mathrm{tor})$ forward.
\item \textbf{Stage 3.} We execute the final rollout to produce the refined pose.
\end{itemize}

\subsubsection{Pose selection}
\label{ssec:pose_selection}

\paragraph{\textsc{GNINA} minimization and scoring.}
For each generated pose, we use \textsc{GNINA}~(v1.3.2) for fast local refinement and scoring.
We run \textsc{GNINA} with \texttt{--minimize} to locally optimize each pose, and output the resulting pose.

For each candidate pose $i$, minimization returns the minimized affinity $E_i$, the heavy-atom RMSD $d_i$ between the generated and minimized structures, and three \textsc{GNINA} CNN outputs: virtual-screening score $v_i$ (\texttt{CNN\_VS}), CNN affinity $a_i$ (\texttt{CNNaffinity}), and pose score $c_i$ (\texttt{CNNscore}). We define the GNINA-based composite score
\begin{equation}
s_i = E_i + d_i - v_i - \tfrac{1}{2}\, a_i - 5\, c_i,
\label{eq:gnina_scoring}
\end{equation}
and retain the pose with the smallest $s_i$. The coefficients are fixed on all benchmarks. Unless stated otherwise, ``scoring'' in what follows refers to this rule.
\textsc{Matcha} is therefore an all-atom model: \textsc{GNINA} is applied at the final stages and operates on explicit protein and ligand atoms, so the output complex is represented at full atomic resolution.

While \textsc{GNINA} is an important step in our method, the analysis in Section~\ref{ssec:pipeline_ablation} demonstrates that replacing Stages 2 and 3 with \textsc{GNINA}-only docking retains only a limited portion of the full performance gain, indicating that the primary source of improvement is the flow-matching generator.

\paragraph{Pose filtration}
We reimplement, speed up, and apply a minimal set of PoseBusters geometric and physicochemical validity filters before scoring.
Specifically, we retain candidate complexes that achieve the highest filter scores across key validity criteria: (i) \emph{Minimum distance to protein}, preventing ligand–receptor atomic collision; (ii) \emph{Protein–ligand maximum distance}, excluding poses with excessive protein–ligand separation that are unlikely to form specific interactions; (iii) \emph{Volume overlap with protein}, rejecting any nonzero volumetric overlap with the receptor; and (iv) \emph{Internal steric clash}, removing ligand conformers with intramolecular clashes.
Filtration is strictly unsupervised (i.e., it does not rely on knowledge of the native pose) and is therefore readily applicable at inference.
In practice, for each complex we retain the poses that pass the highest number of validity criteria.

\subsection{Datasets}
\textsc{Matcha} is trained on two major protein-ligand complex datasets: \textsc{PDBBind}~\citep{liu2017forging} and \textsc{Binding MOAD}~\citep{hu2005binding}.
We follow~\citep{corso2024discovery} in the dataset splitting.
For \textsc{PDBBind}, we use an established time-based split~\citep{stark2022equibind}.
\textsc{Binding MOAD} train split is created so that all \textsc{Binding MOAD} complexes belong to pocket clusters that are present in the \textsc{PDBBind} train set~\citep{corso2024discovery}.
During training, we keep only protein chains close to the ligand (less than $4.5$\,\AA).
We use complexes that have proteins with less than 2000 residues and ligands with $6-150$ heavy atoms.
At training time, we concatenate the \textsc{Binding MOAD} dataset to the \textsc{PDBBind} training set without removing redundant complexes, thereby giving additional weight to higher-quality complexes that passed both datasets' filtering processes.

For the \textsc{Binding MOAD} dataset, we implement protein-level sampling to address the inherent class imbalance where multiple ligands are bound to the same protein structure.
During every training epoch, we sample each unique protein exactly once.
For each selected protein, we randomly choose one ligand from all ligands bound to that receptor.

The docking quality is evaluated on four test datasets.
\textsc{Astex} Diverse set~\citep{hartshorn2007diverse} and \textsc{PoseBusters V2} Benchmark set~\citep{buttenschoen2024posebusters} are commonly used in the field and contain 85 and 308 complexes, respectively.
\textsc{DockGen}~\citep{corso2024discovery} is a set of 330 hard complexes with binding sites different from the training set.
Finally, we use the \textsc{PDBBind} test set, which has 363 complexes deposited in 2019 or later. 

\subsection{Training details}
\label{ssec:train_details}
All models are trained with AdamW using a learning rate of $5\times10^{-5}$ \cite{loshchilov2017decoupled}. The docking models use batch size 24 and are trained on a single NVIDIA H100 80GB GPU. Stage 1 is trained for 1.9M steps, stage 2 for 2.8M steps and stage 3 for 1.4M steps, for a total of approximately 35 GPU-days. All docking models use 12 complex \textsc{DiT} layers with hidden dimension 320, corresponding to roughly 29M parameters. The weighted loss uses coefficients $w_\text{tr}=1$, $w_\text{rot}=1$ and $w_\text{tor}=3$.

\subsection{Docking Quality Metrics}
\label{sec:metrics}

To comprehensively evaluate the performance of \textsc{Matcha}, we use symmetry-corrected Root Mean Square Deviation (RMSD) that accounts for molecular symmetry and report success rates at 2\,\AA\ threshold ($\mathrm{RMSD}\leq\SI{2}{\angstrom}$).
We also run \textsc{PoseBusters} tests~\citep{buttenschoen2024posebusters} to assess physical plausibility of predicted poses, resulting in the combined metric $\mathrm{RMSD}\leq\SI{2}{\angstrom}$ \& PB-valid.
Exact \textsc{PoseBusters} tests are listed in Appendix~\ref{app:posebusters_tests}.
For models that predict whole complex structures, we follow~\cite{abramson2024accurate} and use pocket-aligned symmetric RMSD by aligning the reference protein pocket to the predicted structure.
If a model crashes on a complex, we assign an RMSD of $+\infty$.
The details are reported in Section~\ref{sec:alignment} and Appendix~\ref{app:alignment_strategies}.

\subsection{Docking-tool evaluation protocol}
\label{ssec:benchmark_protocol}
\paragraph{Baseline model parameters}
For \textsc{AlphaFold~3}~\citep{abramson2024accurate}, \textsc{Boltz-2}~\citep{passaro2025boltz} and \textsc{Chai-1}~\citep{chai2024chai}, we use the prediction with the highest confidence score among five model seeds with five samples per seed and 10 recycling steps.
We use multiple sequence alignments computed with \textsc{JackHMMER} (\textsc{HMMER3})~\citep{eddy2011accelerated}.
When modeling receptor structures, we preserve all chains and remove exact duplicate chains only.
\textsc{NeuralPlexer}~\citep{qiao2024state}, \textsc{DiffDock-L}~\citep{corso2024discovery} and \textsc{Uni-Mol}~\citep{alcaide2025uni} are run with their default inference parameters.
We run \textsc{FlowDock}~\citep{morehead2025flowdock}, \textsc{DynamicBind}~\citep{lu2024dynamicbind} and \textsc{SurfDock}~\citep{cao2025surfdock} using true holo protein structures as receptor templates for consistency with other evaluations.
We use the relaxed version of \textsc{DynamicBind}, which is a default behavior.
We detail the parameters used to run classical docking models (\textsc{AutoDock Vina}~(v1.2.5)~\citep{trott2010autodock}, \textsc{SMINA}~(v2020.12.10; fork of Vina~1.1.2)~\citep{koes2013lessons}, and \textsc{GNINA}~(v1.3.2)~\citep{mcnutt2021gnina}) in Appendix~\ref{app:classical_docking_parameters}.
We use the same prepared proteins in \textsc{Matcha} post-processing pipeline as input to \textsc{GNINA}.

\paragraph{Pocket Detection for Classical Docking Methods}
Classical docking methods, \textsc{Uni-Mol} and \textsc{SurfDock} face fundamental challenges in blind docking due to their reliance on pocket identification algorithms.
This dependence introduces additional error when binding sites are unknown, representing a key limitation compared to end-to-end methods.
We evaluate these methods using \textsc{P2Rank}~\citep{krivak2018p2rank}, \textsc{Fpocket}~\citep{le2009fpocket}, and whole protein approaches.
Among these, \textsc{P2Rank} has shown the best performance and is used as the primary pocket detection method for our main results.
The complete results are provided in Appendix~\ref{app:app_results_with_different_pockets}.

\paragraph{\textsc{Matcha} inference parameters}
We run \textsc{Matcha} in two inference regimes: in blind docking setup using all protein chains and in the pocket-aware scenario with the correct binding site provided.
The first docking stage of our approach generates 10 random ligand conformers, which we expand to 20 candidate poses; these are then passed through the refinement stages 2 and 3.
We collect the resulting samples from all three stages and apply \textsc{GNINA} energy minimization, followed by plausibility filtering to discard physically invalid complexes and reduce the candidate set before scoring.
Finally, among the remaining poses, we select the one with the lowest $s_i$ (Eq.~\ref{eq:gnina_scoring}).

We also introduce \textsc{Matcha latte}, a faster variant for settings where speed is prioritized: it uses 5 Euler steps per sample instead of 10, and 10 samples instead of 20.
\textsc{Matcha latte} runs at approximately 8.5\,s per complex on an NVIDIA A100 40GB GPU (see Appendix~\ref{app:timing}), $1.6\times$ faster than the full \textsc{Matcha}, at the cost of a modest drop in docking accuracy.
\textsc{Matcha latte} uses the same three stages and the same post-processing as the full model: Stage~1 generates 10 candidate poses instead of 20, Stages~2 and~3 refine all of them, and the poses that survive validity filtering are ranked with the same scoring rule.

\section*{Funding}
This research received no specific grant from any funding agency in the public, commercial, or not-for-profit sectors.

\section*{Acknowledgements}
We thank Vsevolod Shegolev for the idea of GNINA minimization, and Kamil Garifullin for early experiments with
GNINA integration into the pipeline.

\section*{Author contributions}
DF and TD conducted the experiments; TD carried out the theoretical derivations; ES contributed to the code and derivations.
MP and SN prepared the data; SN, DF and TD ran the baselines. 
TD and DF wrote the manuscript, while DI and MP contributed to editing it.
IO and DI supervised the research.

\section*{Competing interests}
The authors declare no competing interests. 

\section*{Data availability}
This study uses publicly available benchmark datasets: \textsc{PDBBind}~\citep{liu2017forging}, \textsc{Binding MOAD}~\citep{hu2005binding}, the \textsc{Astex} diverse set~\citep{hartshorn2007diverse}, \textsc{PoseBusters V2}~\citep{buttenschoen2024posebusters} and \textsc{DockGen}~\citep{corso2024discovery}, all available from their original sources cited above.
The processed splits, prepared structures and benchmarking artifacts used to reproduce the reported results are available in the \textsc{Matcha} repository at \url{https://github.com/LigandPro/Matcha} and in the code archive at \url{https://doi.org/10.5281/zenodo.23568093}.
No access restrictions apply.

\section*{Code availability}
The \textsc{Matcha} inference and evaluation code is available at \url{https://github.com/LigandPro/Matcha} and archived at \url{https://doi.org/10.5281/zenodo.23568093}.
The model checkpoints used in this study are archived at \url{https://doi.org/10.5281/zenodo.23569055}.

\bibliographystyle{plainnat}
\bibliography{refs}

\begin{thebibliography}{47}
\providecommand{\natexlab}[1]{#1}
\providecommand{\url}[1]{\texttt{#1}}
\expandafter\ifx\csname urlstyle\endcsname\relax
  \providecommand{\doi}[1]{doi: #1}\else
  \providecommand{\doi}{doi: \begingroup \urlstyle{rm}\Url}\fi

\bibitem[Abramson et~al.(2024)Abramson, Adler, Dunger, Evans, Green, Pritzel,
  Ronneberger, Willmore, Ballard, Bambrick, et~al.]{abramson2024accurate}
Josh Abramson, Jonas Adler, Jack Dunger, Richard Evans, Tim Green, Alexander
  Pritzel, Olaf Ronneberger, Lindsay Willmore, Andrew~J Ballard, Joshua
  Bambrick, et~al.
\newblock Accurate structure prediction of biomolecular interactions with
  {AlphaFold} 3.
\newblock \emph{Nature}, 630\penalty0 (8016):\penalty0 493--500, 2024.

\bibitem[Alcaide et~al.(2025)Alcaide, Gao, Ke, Li, Zhang, Zheng, and
  Zhou]{alcaide2025uni}
Eric Alcaide, Zhifeng Gao, Guolin Ke, Yaqi Li, Linfeng Zhang, Hang Zheng, and
  Gengmo Zhou.
\newblock Uni-mol docking v2: Towards realistic and accurate binding pose
  prediction.
\newblock In \emph{International Conference on Artificial Neural Networks},
  pages 34--41. Springer, 2025.

\bibitem[Boitreaud et~al.(2024)Boitreaud, Dent, McPartlon, Meier, Reis,
  Rogozhonikov, and Wu]{chai2024chai}
Jacques Boitreaud, Jack Dent, Matthew McPartlon, Joshua Meier, Vinicius Reis,
  Alex Rogozhonikov, and Kevin Wu.
\newblock Chai-1: Decoding the molecular interactions of life.
\newblock \emph{BioRxiv}, pages 2024--10, 2024.

\bibitem[Br{\'e}gier(2021)]{bregier2021deep}
Romain Br{\'e}gier.
\newblock Deep regression on manifolds: a {3D} rotation case study.
\newblock In \emph{2021 International Conference on 3D Vision (3DV)}, pages
  166--174. IEEE, 2021.

\bibitem[Buttenschoen et~al.(2024)Buttenschoen, Morris, and
  Deane]{buttenschoen2024posebusters}
Martin Buttenschoen, Garrett~M Morris, and Charlotte~M Deane.
\newblock Posebusters: {AI-based docking methods fail to generate physically
  valid poses or generalise to novel sequences}.
\newblock \emph{Chemical Science}, 15\penalty0 (9):\penalty0 3130--3139, 2024.

\bibitem[Cao et~al.(2025)Cao, Chen, Zhang, Wang, Huang, Yu, Jiang, Fan, Zhang,
  Zhou, et~al.]{cao2025surfdock}
Duanhua Cao, Mingan Chen, Runze Zhang, Zhaokun Wang, Manlin Huang, Jie Yu,
  Xinyu Jiang, Zhehuan Fan, Wei Zhang, Hao Zhou, et~al.
\newblock Surfdock is a surface-informed diffusion generative model for
  reliable and accurate protein--ligand complex prediction.
\newblock \emph{Nature Methods}, 22\penalty0 (2):\penalty0 310--322, 2025.

\bibitem[Chen and Lipman(2024)]{chen2024flow}
Ricky~TQ Chen and Yaron Lipman.
\newblock Flow matching on general geometries.
\newblock In \emph{International Conference on Learning Representations},
  volume 2024, pages 47922--47945, 2024.

\bibitem[Chen et~al.(2010)Chen, Arendall~III, Headd, Keedy, Immormino, Kapral,
  Murray, Richardson, and Richardson]{chen2010molprobity}
Vincent~B Chen, W~Bryan Arendall~III, Jeffrey~J Headd, Daniel~A Keedy, Robert~M
  Immormino, Gary~J Kapral, Laura~W Murray, Jane~S Richardson, and David~C
  Richardson.
\newblock Molprobity: all-atom structure validation for macromolecular
  crystallography.
\newblock \emph{Acta Crystallographica Section D: Biological Crystallography},
  66\penalty0 (1):\penalty0 12--21, 2010.
\newblock \doi{10.1107/S0907444909042073}.

\bibitem[Corso et~al.(2023)Corso, Stärk, Jing, Barzilay, and
  Jaakkola]{corso2023diffdock}
Gabriele Corso, Hannes Stärk, Bowen Jing, Regina Barzilay, and Tommi Jaakkola.
\newblock {DiffDock: Diffusion Steps, Twists, and Turns for Molecular Docking}.
\newblock In \emph{International Conference on Learning Representations
  (ICLR)}, 2023.

\bibitem[Corso et~al.(2024)Corso, Deng, Polizzi, Barzilay, and
  Jaakkola]{corso2024discovery}
Gabriele Corso, Arthur Deng, Nicholas Polizzi, Regina Barzilay, and Tommi
  Jaakkola.
\newblock Deep confident steps to new pockets: Strategies for docking
  generalization.
\newblock In \emph{International Conference on Learning Representations
  (ICLR)}, 2024.

\bibitem[DeLano et~al.(2002)]{delano2002pymol}
Warren~L DeLano et~al.
\newblock Pymol: An open-source molecular graphics tool.
\newblock \emph{CCP4 Newsl. protein crystallogr}, 40\penalty0 (1):\penalty0
  82--92, 2002.

\bibitem[Eddy(2011)]{eddy2011accelerated}
Sean~R Eddy.
\newblock Accelerated profile {HMM} searches.
\newblock \emph{PLoS Computational Biology}, 7\penalty0 (10):\penalty0
  e1002195, 2011.

\bibitem[Forli et~al.(2016)Forli, Huey, Pique, Sanner, Goodsell, and
  Olson]{forli2016computational}
Stefano Forli, Ruth Huey, Michael~E Pique, Michel~F Sanner, David~S Goodsell,
  and Arthur~J Olson.
\newblock {Computational protein--ligand docking and virtual drug screening
  with the AutoDock suite}.
\newblock \emph{{Nature Protocols}}, 11\penalty0 (5):\penalty0 905--919, 2016.

\bibitem[{Forli Lab, CCSB} et~al.(){Forli Lab, CCSB}, Santos-Martins,
  Eberhardt, and Forli]{Meeko}
{Forli Lab, CCSB}, D.~Santos-Martins, J.~Eberhardt, and S.~Forli.
\newblock Meeko: Interface for {AutoDock} and {RDKit}.
\newblock \url{https://meeko.readthedocs.io/} and
  \url{https://github.com/forlilab/Meeko}.
\newblock Version 0.4.0, accessed 2025-09-11.

\bibitem[Friesner et~al.(2004)Friesner, Banks, Murphy, Halgren, Klicic, Mainz,
  Repasky, Knoll, Shelley, Perry, et~al.]{friesner2004glide}
Richard~A Friesner, Jay~L Banks, Robert~B Murphy, Thomas~A Halgren, Jasna~J
  Klicic, Daniel~T Mainz, Matthew~P Repasky, Eric~H Knoll, Mee Shelley, Jason~K
  Perry, et~al.
\newblock {Glide: a new approach for rapid, accurate docking and scoring. 1.
  Method and assessment of docking accuracy}.
\newblock \emph{{Journal of Medicinal Chemistry}}, 47\penalty0 (7):\penalty0
  1739--1749, 2004.

\bibitem[Hartshorn et~al.(2007)Hartshorn, Verdonk, Chessari, Brewerton, Mooij,
  Mortenson, and Murray]{hartshorn2007diverse}
Michael~J Hartshorn, Marcel~L Verdonk, Gianni Chessari, Suzanne~C Brewerton,
  Wijnand~TM Mooij, Paul~N Mortenson, and Christopher~W Murray.
\newblock Diverse, high-quality test set for the validation of protein-ligand
  docking performance.
\newblock \emph{{Journal of Medicinal Chemistry}}, 50\penalty0 (4):\penalty0
  726--741, 2007.

\bibitem[Hu et~al.(2005)Hu, Benson, Smith, Lerner, and Carlson]{hu2005binding}
Liegi Hu, Mark~L Benson, Richard~D Smith, Michael~G Lerner, and Heather~A
  Carlson.
\newblock Binding {MOAD} (mother of all databases).
\newblock \emph{Proteins: Structure, Function, and Bioinformatics}, 60\penalty0
  (3):\penalty0 333--340, 2005.

\bibitem[Jumper et~al.(2021)Jumper, Evans, Pritzel, Green, Figurnov,
  Ronneberger, Tunyasuvunakool, Bates, {\v{Z}}{\'\i}dek, Potapenko,
  et~al.]{jumper2021highly}
John Jumper, Richard Evans, Alexander Pritzel, Tim Green, Michael Figurnov,
  Olaf Ronneberger, Kathryn Tunyasuvunakool, Russ Bates, Augustin
  {\v{Z}}{\'\i}dek, Anna Potapenko, et~al.
\newblock {Highly accurate protein structure prediction with AlphaFold}.
\newblock \emph{{Nature}}, 596\penalty0 (7873):\penalty0 583--589, 2021.

\bibitem[Koes et~al.(2013)Koes, Baumgartner, and Camacho]{koes2013lessons}
David~Ryan Koes, Matthew~P Baumgartner, and Carlos~J Camacho.
\newblock Lessons learned in empirical scoring with {SMINA} from the {CSAR}
  2011 benchmarking exercise.
\newblock \emph{{Journal of Chemical Information and Modeling}}, 53\penalty0
  (8):\penalty0 1893--1904, 2013.

\bibitem[Kriv{\'a}k and Hoksza(2018)]{krivak2018p2rank}
Radoslav Kriv{\'a}k and David Hoksza.
\newblock {P2Rank}: machine learning based tool for rapid and accurate
  prediction of ligand binding sites from protein structure.
\newblock \emph{Journal of Cheminformatics}, 10\penalty0 (1):\penalty0 39,
  2018.

\bibitem[Landrum(2024)]{landrum2024rdkit}
Greg Landrum.
\newblock {RDKit: Open-Source Cheminformatics Software}, 2024.
\newblock URL
  \url{https://github.com/rdkit/rdkit/releases/tag/Release_2024_09_1}.
\newblock Release 2024\_09\_1.

\bibitem[Le~Guilloux et~al.(2009)Le~Guilloux, Schmidtke, and
  Tuffery]{le2009fpocket}
Vincent Le~Guilloux, Peter Schmidtke, and Pierre Tuffery.
\newblock Fpocket: an open source platform for ligand pocket detection.
\newblock \emph{BMC Bioinformatics}, 10\penalty0 (1):\penalty0 168, 2009.

\bibitem[Lin et~al.(2022)Lin, Akin, Rao, Hie, Zhu, Lu, dos Santos~Costa,
  Fazel-Zarandi, Sercu, Candido, et~al.]{lin2022language}
Zeming Lin, Halil Akin, Roshan Rao, Brian Hie, Zhongkai Zhu, Wenting Lu, Allan
  dos Santos~Costa, Maryam Fazel-Zarandi, Tom Sercu, Sal Candido, et~al.
\newblock {Language Models of Protein Sequences at the Scale of Evolution
  Enable Accurate Structure Prediction}.
\newblock \emph{BioRxiv}, 2022:\penalty0 500902, 2022.

\bibitem[Lipinski and Hopkins(2004)]{lipinski2004navigating}
Christopher Lipinski and Andrew Hopkins.
\newblock Navigating chemical space for biology and medicine.
\newblock \emph{Nature}, 432\penalty0 (7019):\penalty0 855--861, 2004.

\bibitem[Lipman et~al.(2023)Lipman, Chen, Ben-Hamu, Nickel, and
  Le]{lipman2023flow}
Yaron Lipman, Ricky T.~Q. Chen, Heli Ben-Hamu, Maximilian Nickel, and Matthew
  Le.
\newblock Flow matching for generative modeling.
\newblock In \emph{The Eleventh International Conference on Learning
  Representations}, 2023.
\newblock URL \url{https://openreview.net/forum?id=PqvMRDCJT9t}.

\bibitem[Liu et~al.(2015)Liu, Li, Han, Li, Liu, Zhao, Nie, Liu, and
  Wang]{liu2015pdb}
Zhihai Liu, Yan Li, Li~Han, Jie Li, Jie Liu, Zhixiong Zhao, Wei Nie, Yuchen
  Liu, and Renxiao Wang.
\newblock Pdb-wide collection of binding data: current status of the pdbbind
  database.
\newblock \emph{Bioinformatics}, 31\penalty0 (3):\penalty0 405--412, 2015.

\bibitem[Liu et~al.(2017)Liu, Su, Han, Liu, Yang, Li, and Wang]{liu2017forging}
Zhihai Liu, Minyi Su, Li~Han, Jie Liu, Qifan Yang, Yan Li, and Renxiao Wang.
\newblock Forging the basis for developing protein--ligand interaction scoring
  functions.
\newblock \emph{Accounts of chemical research}, 50\penalty0 (2):\penalty0
  302--309, 2017.

\bibitem[Loshchilov and Hutter(2017)]{loshchilov2017decoupled}
Ilya Loshchilov and Frank Hutter.
\newblock {Decoupled Weight Decay Regularization}.
\newblock \emph{arXiv preprint arXiv:1711.05101}, 2017.

\bibitem[Lu et~al.(2024)Lu, Zhang, Huang, Zhang, Jia, Wang, Shi, Li, Wolynes,
  and Zheng]{lu2024dynamicbind}
Wei Lu, Jixian Zhang, Weifeng Huang, Ziqiao Zhang, Xiangyu Jia, Zhenyu Wang,
  Leilei Shi, Chengtao Li, Peter~G Wolynes, and Shuangjia Zheng.
\newblock {DynamicBind}: predicting ligand-specific protein-ligand complex
  structure with a deep equivariant generative model.
\newblock \emph{Nature Communications}, 15\penalty0 (1):\penalty0 1071, 2024.

\bibitem[McNutt et~al.(2021)McNutt, Francoeur, Aggarwal, Masuda, Meli, Ragoza,
  Sunseri, and Koes]{mcnutt2021gnina}
Andrew~T McNutt, Paul Francoeur, Rishal Aggarwal, Tomohide Masuda, Rocco Meli,
  Matthew Ragoza, Jocelyn Sunseri, and David~Ryan Koes.
\newblock Gnina 1.0: molecular docking with deep learning.
\newblock \emph{{Journal of Cheminformatics}}, 13\penalty0 (1):\penalty0 43,
  2021.

\bibitem[Morehead and Cheng(2025)]{morehead2025flowdock}
Alex Morehead and Jianlin Cheng.
\newblock Flowdock: Geometric flow matching for generative protein--ligand
  docking and affinity prediction.
\newblock \emph{Bioinformatics}, 41, 2025.

\bibitem[Morehead et~al.(2026)Morehead, Giri, Liu, Neupane, and
  Cheng]{morehead2026assessing}
Alex Morehead, Nabin Giri, Jian Liu, Pawan Neupane, and Jianlin Cheng.
\newblock Assessing the potential of deep learning for protein--ligand docking.
\newblock \emph{Nature Machine Intelligence}, 8\penalty0 (1):\penalty0 32--41,
  2026.

\bibitem[Pak et~al.(2025)Pak, Frolova, Nikolenko, Daulbaev, Ryabchenko, Litvin,
  Gurevich, Garifullin, Shapeev, Oseledets, et~al.]{pak2025bento}
Marina~A Pak, Daria Frolova, Sergei~A Nikolenko, Talgat Daulbaev, Daria
  Ryabchenko, Anna Litvin, Pavel Gurevich, Kamil Garifullin, Alexander Shapeev,
  Ivan Oseledets, et~al.
\newblock Bento: Benchmarking classical and ai docking on drug design-relevant
  data.
\newblock \emph{BioRxiv}, pages 2025--12, 2025.

\bibitem[Passaro et~al.(2025)Passaro, Corso, Wohlwend, Reveiz, Thaler, Somnath,
  Getz, Portnoi, Roy, Stark, et~al.]{passaro2025boltz}
Saro Passaro, Gabriele Corso, Jeremy Wohlwend, Mateo Reveiz, Stephan Thaler,
  Vignesh~Ram Somnath, Noah Getz, Tally Portnoi, Julien Roy, Hannes Stark,
  et~al.
\newblock Boltz-2: Towards accurate and efficient binding affinity prediction.
\newblock \emph{BioRxiv}, 2025.

\bibitem[Peebles and Xie(2023)]{peebles2023scalable}
William Peebles and Saining Xie.
\newblock Scalable diffusion models with transformers.
\newblock In \emph{Proceedings of the IEEE/CVF International Conference on
  Computer Vision}, pages 4195--4205, 2023.

\bibitem[Qiao et~al.(2026)Qiao, Ding, Dresselhaus, Rosenfeld, Han, Howell,
  Iyengar, Opalenski, Christensen, Sirumalla, et~al.]{qiao2026neuralplexer3}
Jarren~Zhuoran Qiao, Feizhi Ding, Thomas Dresselhaus, Mia Rosenfeld, Xiaotian
  Han, Owen Howell, Aniketh Iyengar, Stephen Opalenski, Anders Christensen,
  Sai~Krishna Sirumalla, et~al.
\newblock Neuralplexer3: accurate biomolecular complex structure prediction
  with flow models.
\newblock \emph{Advances in Neural Information Processing Systems},
  38:\penalty0 1--38, 2026.

\bibitem[Qiao et~al.(2024)Qiao, Nie, Vahdat, Miller~III, and
  Anandkumar]{qiao2024state}
Zhuoran Qiao, Weili Nie, Arash Vahdat, Thomas~F Miller~III, and Animashree
  Anandkumar.
\newblock State-specific protein--ligand complex structure prediction with a
  multiscale deep generative model.
\newblock \emph{Nature Machine Intelligence}, 6\penalty0 (2):\penalty0
  195--208, 2024.

\bibitem[Riniker and Landrum(2015)]{riniker2015etkdg}
Sereina Riniker and Gregory~A Landrum.
\newblock Better informed distance geometry: using what we know to improve
  conformer generation.
\newblock \emph{Journal of Chemical Information and Modeling}, 55\penalty0
  (12):\penalty0 2562--2574, 2015.
\newblock \doi{10.1021/acs.jcim.5b00654}.

\bibitem[Sadybekov and Katritch(2023)]{sadybekov2023computational}
Anastasiia~V Sadybekov and Vsevolod Katritch.
\newblock Computational approaches streamlining drug discovery.
\newblock \emph{Nature}, 616\penalty0 (7958):\penalty0 673--685, 2023.

\bibitem[Shoemake(1985)]{shoemake1985animating}
Ken Shoemake.
\newblock Animating rotation with quaternion curves.
\newblock In \emph{Proceedings of the 12th annual conference on Computer
  Graphics and Interactive Techniques}, pages 245--254, 1985.

\bibitem[St{\"a}rk et~al.(2022)St{\"a}rk, Ganea, Pattanaik, Barzilay, and
  Jaakkola]{stark2022equibind}
Hannes St{\"a}rk, Octavian Ganea, Lagnajit Pattanaik, Regina Barzilay, and
  Tommi Jaakkola.
\newblock Equibind: Geometric deep learning for drug binding structure
  prediction.
\newblock In \emph{International Conference on Machine Learning}, pages
  20503--20521. PMLR, 2022.

\bibitem[Tropsha et~al.(2024)Tropsha, Isayev, Varnek, Schneider, and
  Cherkasov]{tropsha2024integrating}
Alexander Tropsha, Olexandr Isayev, Alexandre Varnek, Gisbert Schneider, and
  Artem Cherkasov.
\newblock {Integrating QSAR modelling and deep learning in drug discovery: the
  emergence of deep QSAR}.
\newblock \emph{Nature Reviews Drug Discovery}, 23\penalty0 (2):\penalty0
  141--155, 2024.

\bibitem[Trott and Olson(2010)]{trott2010autodock}
Oleg Trott and Arthur~J Olson.
\newblock {AutoDock Vina}: improving the speed and accuracy of docking with a
  new scoring function, efficient optimization, and multithreading.
\newblock \emph{{Journal of Computational Chemistry}}, 31\penalty0
  (2):\penalty0 455--461, 2010.

\bibitem[Warner(1983)]{warner1983foundations}
Frank~W Warner.
\newblock \emph{Foundations of differentiable manifolds and Lie groups},
  volume~94.
\newblock Springer Science \& Business Media, 1983.

\bibitem[Wohlwend et~al.(2025)Wohlwend, Corso, Passaro, Getz, Reveiz, Leidal,
  Swiderski, Atkinson, Portnoi, Chinn, et~al.]{wohlwend2025boltz}
Jeremy Wohlwend, Gabriele Corso, Saro Passaro, Noah Getz, Mateo Reveiz, Ken
  Leidal, Wojtek Swiderski, Liam Atkinson, Tally Portnoi, Itamar Chinn, et~al.
\newblock Boltz-1 democratizing biomolecular interaction modeling.
\newblock \emph{BioRxiv}, pages 2024--11, 2025.

\bibitem[Word et~al.(1999)Word, Lovell, Richardson, and
  Richardson]{Word1999AsnGln}
J.~M. Word, S.~C. Lovell, J.~S. Richardson, and D.~C. Richardson.
\newblock Asparagine and glutamine: using hydrogen atom contacts in the choice
  of side-chain amide orientation.
\newblock \emph{Journal of Molecular Biology}, 285\penalty0 (4):\penalty0
  1735--1747, 1999.
\newblock \doi{10.1006/jmbi.1998.2401}.

\bibitem[Zhou et~al.(2023)Zhou, Gao, Ding, Zheng, Xu, Wei, Zhang, and
  Ke]{zhou2023uni}
Gengmo Zhou, Zhifeng Gao, Qiankun Ding, Hang Zheng, Hongteng Xu, Zhewei Wei,
  Linfeng Zhang, and Guolin Ke.
\newblock {Uni-Mol}: A universal {3D} molecular representation learning
  framework.
\newblock In \emph{The Eleventh International Conference on Learning
  Representations}, 2023.
\newblock URL \url{https://openreview.net/forum?id=6K2RM6wVqKu}.

\end{thebibliography}

\clearpage
\begin{appendices}
\section{Flow matching on $SO(2)$ and $SO(3)$}
\label{sec:app_so_n}
In this section, we address the problem of training flow matching on the special orthogonal groups $SO(2)$ and $SO(3)$. Our approach involves minimizing the flow matching loss function with respect to the parameters $\vw$ of a flow model denoted by $v$. The loss function is defined as:

\begin{equation}
\ell(\vw) = \E_{\vx_0, \vx_1, t} \left\| v(\vx(t), t; \vw) - \frac{\mathrm{d} \vx(t)}{\mathrm{d} t} \right\|_g^2,
\label{eq::sup_loss}
\end{equation}

In the equation above, $\vx(t)$ represents an interpolation between two points $\vx_0$ and $\vx_1$.
The term $\| \vx \|_g = \sqrt{g(\vx, \vx)}$ defines a norm based on the Riemannian metric $g$.

For a given Riemannian manifold \(M\), the tangent space at a point \(p \in M\) is denoted by \(\mathcal{T}_p M\). The Riemannian metric is defined as:

\begin{equation}
    g: \mathcal{T}_p M \times \mathcal{T}_p M \to \mathbb{R},
\end{equation}

acts on the Cartesian product of these tangent spaces to produce a non-negative scalar value.

The special orthogonal group \(SO(n)\) consists of elements that can be represented as \(n \times n\) rotation matrices.
Although there are several ways to define a Riemannian metric \(g\) on \(SO(n)\), the canonical metric is given by:
\begin{equation}
    g(\mathbf{X}, \mathbf{Y}) \triangleq \mathrm{tr} (\mathbf{X}^\top \mathbf{Y}),
\label{eq::riemannian_metric}
\end{equation}

where \(\mathbf{X}\) and \(\mathbf{Y}\) are \(n \times n\) matrices corresponding to the elements of the tangent space.

Regarding interpolation on these manifolds, various methods exist.
In this work, we employ the widely used spherical linear interpolation, commonly referred to as SLERP~\cite{shoemake1985animating}.

\subsection{$SO(2)$ manifold}

Every element of $\mathrm{SO}(2)$ can be parameterized by an angle $\theta$ and written as
\[
R(\theta)=
\begin{bmatrix}
\cos\theta & -\sin\theta \\
\sin\theta & \cos\theta
\end{bmatrix}.
\]
Since angles on $\mathrm{SO}(2)$ are periodic, the interpolation between two angles should not
be defined using the raw difference $\theta_1-\theta_0$. Instead, we use the signed shortest
angular displacement
\[
\Delta\theta =
\operatorname{wrap}_{(-\pi,\pi]}(\theta_1-\theta_0),
\]
where $\operatorname{wrap}_{(-\pi,\pi]}$ maps an angle to the interval $(-\pi,\pi]$. The
corresponding geodesic interpolation is
\[
\theta(t)=\theta_0+t\Delta\theta, \qquad t\in[0,1].
\]
This choice follows the shortest path on the circle and avoids discontinuities at the branch cut
$\pm\pi$.

The time derivative of the interpolated rotation matrix is
\[
\dot R(t)=
\begin{bmatrix}
-\sin\theta(t) & -\cos\theta(t) \\
 \cos\theta(t) & -\sin\theta(t)
\end{bmatrix}
\Delta\theta.
\]
Equivalently,
\[
\dot R(t)=
R(\theta(t))
\begin{bmatrix}
0 & -\Delta\theta \\
\Delta\theta & 0
\end{bmatrix}.
\]
Thus, the tangent vector can be represented by the scalar angular velocity $\Delta\theta$.

Assume that the neural network predicts a scalar velocity
$\hat v(\theta(t),t;w)$. The corresponding tangent vector is
\[
v(R(t),t;w)=
R(\theta(t))
\begin{bmatrix}
0 & -\hat v(\theta(t),t;w) \\
\hat v(\theta(t),t;w) & 0
\end{bmatrix}.
\]
Using the canonical metric
\[
g(X,Y)=\operatorname{tr}(X^\top Y),
\]
we obtain
\[
\|v(R(t),t;w)-\dot R(t)\|_g^2
=
2\left(\hat v(\theta(t),t;w)-\Delta\theta\right)^2.
\]
The constant factor does not affect the minimizer, so in practice we minimize the equivalent
squared loss
\[
\ell(w)=
\mathbb{E}_{\theta_0,\theta_1,t}
\left(
\hat v(\theta(t),t;w)-\Delta\theta
\right)^2.
\]

For torsional angles with a symmetry period $p$, we use the same construction with the wrapped
shortest displacement
\[
\Delta\theta_p =
\operatorname{wrap}_{(-p/2,p/2]}(\theta_1-\theta_0),
\]
and set
\[
\theta(t)=\theta_0+t\Delta\theta_p.
\]
The standard $\mathrm{SO}(2)$ case above corresponds to $p=2\pi$.

\subsection{$SO(3)$ manifold}

In contrast to $SO(2)$ case, SLERP for $3 \times 3$ rotation matrices is more complicated, often requiring a transformation into quaternions~\citep{bregier2021deep}.
Nonetheless, it is possible to compute the interpolated matrix $\mathbf{R}(t)$ and its corresponding time derivative, which belongs to the tangent space of $\mathbf{R}(t)$, using automatic differentiation tools.

Every tangent vector of a point $\mathbf{R} \in SO(3)$ can be expressed as
\begin{equation}
\mathbf{R}
\begin{bmatrix}
0 & -k_z & k_y \\
k_z & 0 & -k_x \\
-k_y & k_x & 0 \\
\end{bmatrix}.
\end{equation}

Assuming that the values $k_x, k_y, k_z$ describe the matrix $\dot{\mathbf{R}}(t)$, a neural network can be constructed to yield three outputs: $v_x, v_y, v_z$. The flow model becomes
\begin{equation}
    v(\mathbf{R}(t), t) = \mathbf{R}(t)
    \begin{bmatrix}
        0 & -v_z & v_y \\
        v_z & 0 & -v_x \\
        -v_y & v_x & 0 \\
    \end{bmatrix}.
\end{equation}

For the $SO(3)$ manifold, the square of norm in equation~(\ref{eq::sup_loss}) evaluates to
\begin{equation}
2(k_x - v_x)^2 + 2(k_y - v_y)^2 + 2(k_z - v_z)^2.
\end{equation}

To determine the values $k_x, k_y, k_z$, one can leverage automatic differentiation to compute $\dot{\mathbf{R}}(t)$. Subsequently, these values can be extracted from the expression $\mathbf{R}^\top(t) \dot{\mathbf{R}}(t)$.

\section{Details on running classical docking}
\label{app:classical_docking_parameters}
\paragraph{Classical docking baseline parameters}\mbox{}\\[0.25\baselineskip]
\noindent
We used \textsc{AutoDock Vina}~(v1.2.5)~\citep{trott2010autodock},
\textsc{SMINA}~(v2020.12.10; fork of Vina~1.1.2)~\citep{koes2013lessons},\newline
and~\mbox{\textsc{GNINA}}~(v1.3.2)~\citep{mcnutt2021gnina}.
Unless noted otherwise, all runs used \texttt{exhaustiveness 64} and \texttt{seed 42}.
Pocket centers were provided as described below; the search box was centered at each pocket center and sized to the RDKit conformer diameter of the ligand plus a 10~\AA\ padding on all six sides (equivalently, \texttt{autobox\_add 10} where applicable).
To emulate blind docking we also used a large box centered on the protein with padding \texttt{autobox\_add 16}.

\paragraph{Ligand and receptor preparation for classical docking}
Each ligand was prepared from a SMILES string: standardization and neutralization of charges, adjustment to pH~7 protonation rules, addition of explicit hydrogens, and 3D conformer generation with RDKit’s ETKDG method~\citep{riniker2015etkdg}.
Receptor proteins were hydrogenated with \textsc{Reduce}~(v4.13) using the \texttt{FLIP} option (Asparagine (Asn), Glutamine (Gln), Histidine (His) side-chain flips)~\citep{Word1999AsnGln,chen2010molprobity}.
For AutoDock-family tools, inputs were converted to PDBQT with \textsc{Meeko}~(v0.4.0)~\citep{Meeko}.

\section{Inference speed comparison}
\label{app:timing}

\begin{table}[htb]
  \caption{Comparison of the average inference time for blind docking models (for \textsc{Astex} test set)}
  \label{tab:inference_speed}
  \centering
  \small
  \begin{tabular}{@{}p{0.70\columnwidth}r@{}}
    \toprule
    Method & Time (s) \\
    \midrule
    \textsc{Matcha latte} & 8.5 \\
    \textsc{Matcha} & 13.6 \\
    \textsc{DiffDock-L} & 32 \\
    \textsc{NeuralPlexer} & 65\\
    \textsc{FlowDock} & 39 \\
    \textsc{GNINA} (P2Rank) for one initial pose & 18 \\
    \textsc{AlphaFold} 3 & 392 \\
    \textsc{Chai-1} & 1638\\
    \textsc{Boltz-2} & 1488\\
    \bottomrule
  \end{tabular}
\end{table}
We report the average inference speed for all considered blind docking methods on one NVIDIA A100 40GB GPU.
Timing is measured on the \textsc{Astex} set ($n=85$).
Per-complex runtime is a property of the method and the hardware, so this small diverse set is enough to estimate it.
The success rates in the main-text timing figure are instead taken from \textsc{PoseBusters V2}, the larger held-out benchmark used for that accuracy comparison.
Time is reported only for model inference avoiding model loading.
Most docking models generate multiple poses and select a pose with the best score, so we measure the time required to sample all required poses.
\textsc{Matcha latte} is our lighter variant (5 Euler steps, 10 samples); it achieves the fastest inference among the reported methods at 8.5\,s per complex.
At the same time, \textsc{Matcha} takes 13.6\,s to predict one complex on average.
The results are shown in Table~\ref{tab:inference_speed}.

Additionally, we measured time required to perform \textsc{Matcha} stages: docking (pose generation) takes 7.35\,s while \textsc{GNINA} minimization, validity filtering and scoring take 6.21\,s per complex.

\section{Blind docking results with different pocket prediction methods}
\label{app:app_results_with_different_pockets}
In this section, we report results for blind docking scenario for models that require pocket information as input: \textsc{Uni-Mol}, \textsc{SurfDock}, \textsc{SMINA}, \textsc{VINA}, \textsc{GNINA}.
We use three types of pocket information: \textsc{P2Rank}~\citep{krivak2018p2rank}, \textsc{Fpocket}~\citep{le2009fpocket}, and the full protein setup.
Full protein means providing the whole protein and the protein center as a starting pocket center: using large box centered on the protein with padding \texttt{autobox\_add 16}.
The full protein setup was not used for \textsc{Uni-Mol} and \textsc{SurfDock} because these models are unable to process such input.

\begin{figure*}[!t]
  \centering
  \includegraphics[width=\textwidth]{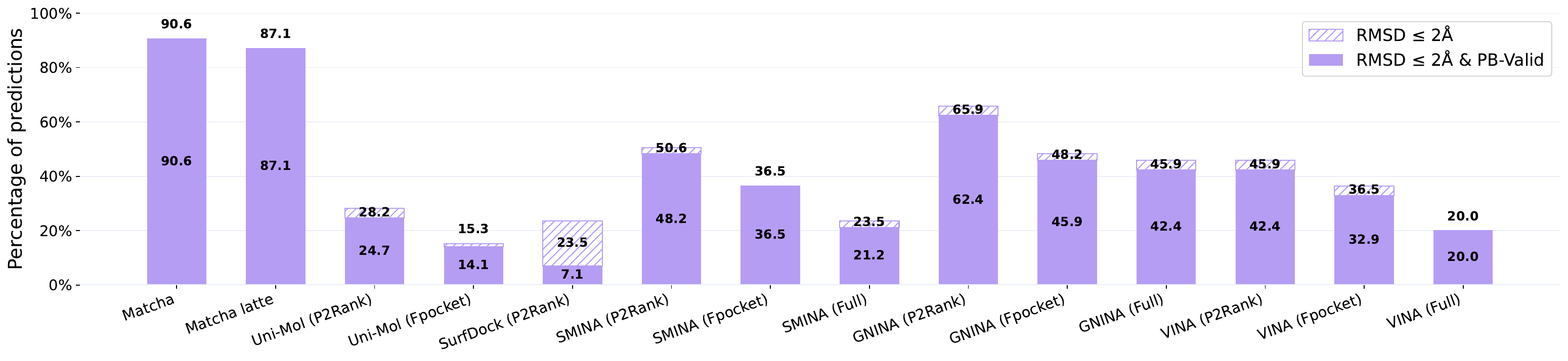}
\caption{Blind ligand docking success rates with different pocket prediction methods on \textsc{Astex} Diverse set.}
\label{fig:res_astex_fpocket}

\end{figure*}

\begin{figure*}[!t]
  \centering
  \includegraphics[width=\textwidth]{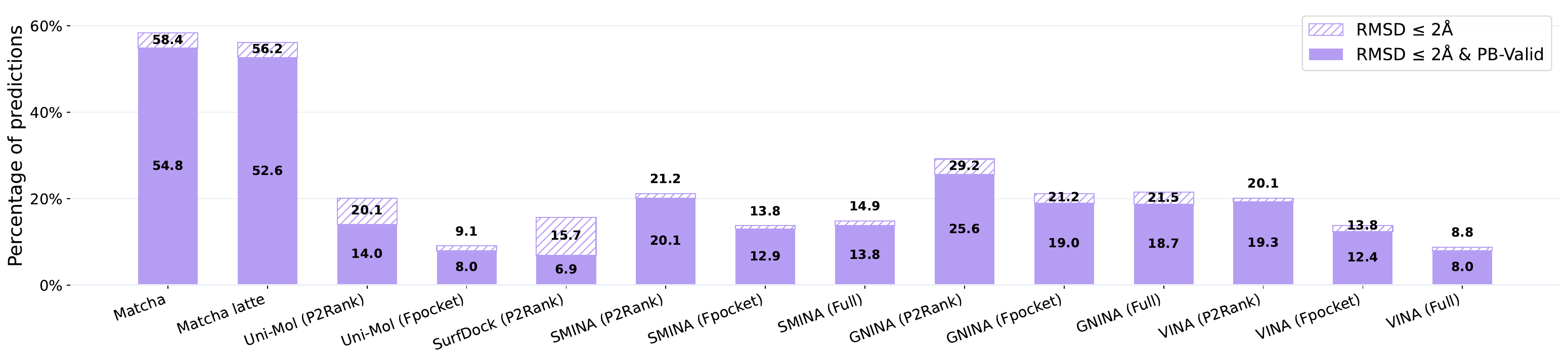}
\caption{Blind ligand docking success rates with different pocket prediction methods on \textsc{PDBBind} test set.}
\label{fig:res_pdbbind_fpocket}

\end{figure*}

\begin{figure*}[!t]
  \centering
  \includegraphics[width=\textwidth]{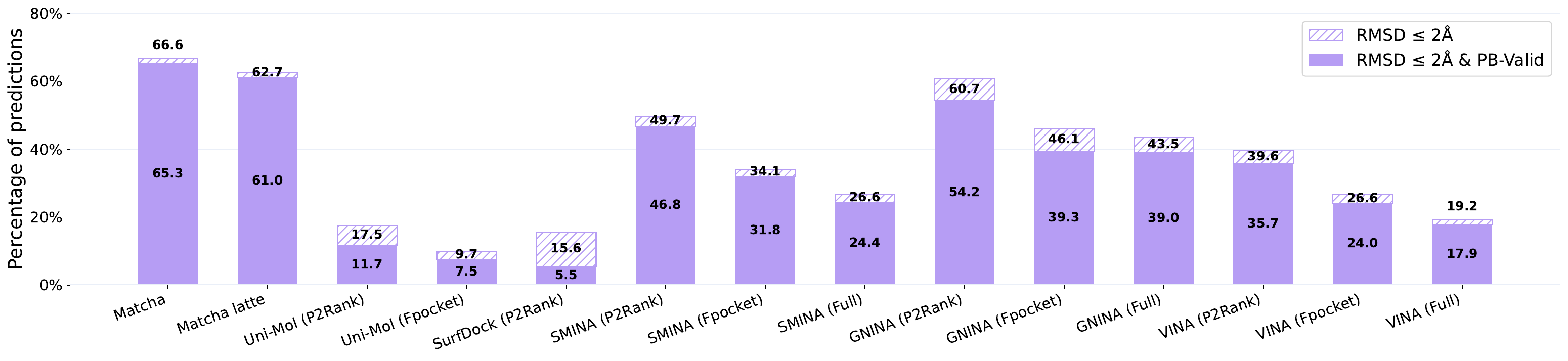}
\caption{Blind ligand docking success rates with different pocket prediction methods on \textsc{PoseBusters V2} dataset.}
\label{fig:res_posebusters_fpocket}

\end{figure*}

\begin{figure*}[!t]
  \centering
  \includegraphics[width=\textwidth]{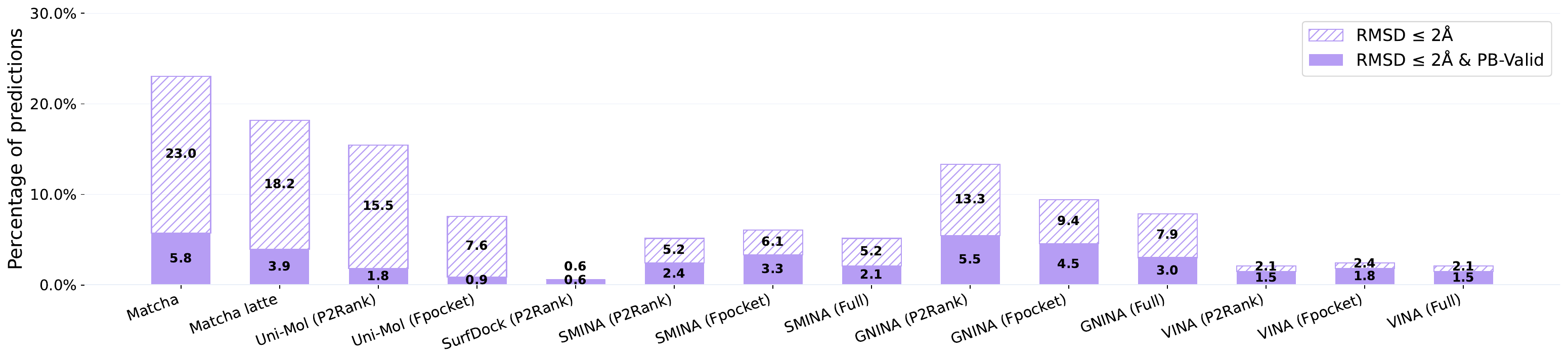}
\caption{Blind ligand docking success rates with different pocket prediction methods on \textsc{DockGen} dataset.}
\label{fig:res_dockgen_fpocket}

\end{figure*}

We report the results for all four considered test datasets in Figures~\ref{fig:res_astex_fpocket}, \ref{fig:res_pdbbind_fpocket}, \ref{fig:res_posebusters_fpocket}, and \ref{fig:res_dockgen_fpocket}.
According to them, \textsc{P2Rank} consistently shows the best quality among all other pocket identification strategies.

\section{The importance of the pocket alignment}
\label{app:alignment}

\subsection{Pocket-aligned RMSD computation}
\label{app:alignment_strategies}

For all models that predict the structure of the whole complex, docking RMSD depends on how the predicted and reference proteins are aligned before RMSD is computed. Recent co-folding studies are not fully consistent in this choice: \textsc{AlphaFold 3}~\citep{abramson2024accurate} uses a \textsc{full} pocket alignment, whereas \textsc{NeuralPlexer 3}~\citep{qiao2026neuralplexer3} uses a \textsc{pocket-based} alignment. In the main text, we report pocket-aligned symmetric RMSD with the \textsc{pocket-based} alignment for direct comparability with \textsc{NeuralPlexer 3}-style reporting. In this section, we define both procedures and use the \textsc{full} alignment as a robustness check, which also facilitates comparison to the \textsc{AlphaFold 3} protocol.
We first describe our \textsc{full} alignment procedure.

\begin{enumerate}
    \item The primary protein chain with the most atoms within 10\,\AA\ of the ligand is kept.
    \item The pocket is defined as all C$_\alpha$ atoms within 10\,\AA\ of any heavy atom of the reference ligand, restricted to protein backbone atoms.
    \item The reference pocket is aligned to the \textit{whole predicted protein structure} by C$_\alpha$ atoms in PyMOL~\citep{delano2002pymol} with \textit{five refinement cycles}, which is the default parameter.
\end{enumerate}

An alternative approach to compute pocket-aware RMSD was described in~\citet{qiao2026neuralplexer3}.
This \textsc{pocket-based} approach shares the first two stages, but then the procedure differs:

\begin{enumerate}
\addtocounter{enumi}{2}
    \item The predicted pocket is defined as all C$_\alpha$ atoms within 10\,\AA\ of any heavy atom of the predicted ligand, restricted to protein backbone atoms.
    \item \textit{Each chain in the predicted pocket} is aligned to the reference pocket, and the chain with the minimum alignment RMSD is selected.
    The alignment is performed with \textit{zero refinement cycles}.
\end{enumerate}

The \textsc{pocket-based} alignment is useful for comparison with prior work, but in blind docking it can yield more optimistic results because it reduces the penalty for pocket-localization errors. This is particularly relevant for multi-chain proteins or proteins with several similar binding sites, where an incorrect pocket may align well by chance. By contrast, the \textsc{full} alignment penalizes such failures more directly because the reference pocket is aligned to the entire predicted protein structure rather than to a predicted local pocket.

This difference in alignment protocol explains why our metrics differ from the \textsc{AlphaFold 3} results reported in the \textsc{NeuralPlexer 3} paper on the \textsc{PoseBusters V2} dataset. A comparison between the \textsc{full} and \textsc{pocket-based} approaches is provided in Appendix~\ref{app:aligned_results}.

\subsection{Docking results for different ways of computing pocket-aligned RMSD}
\label{app:aligned_results}

\begin{figure*}[!t]
  \centering
  \includegraphics[width=\textwidth]{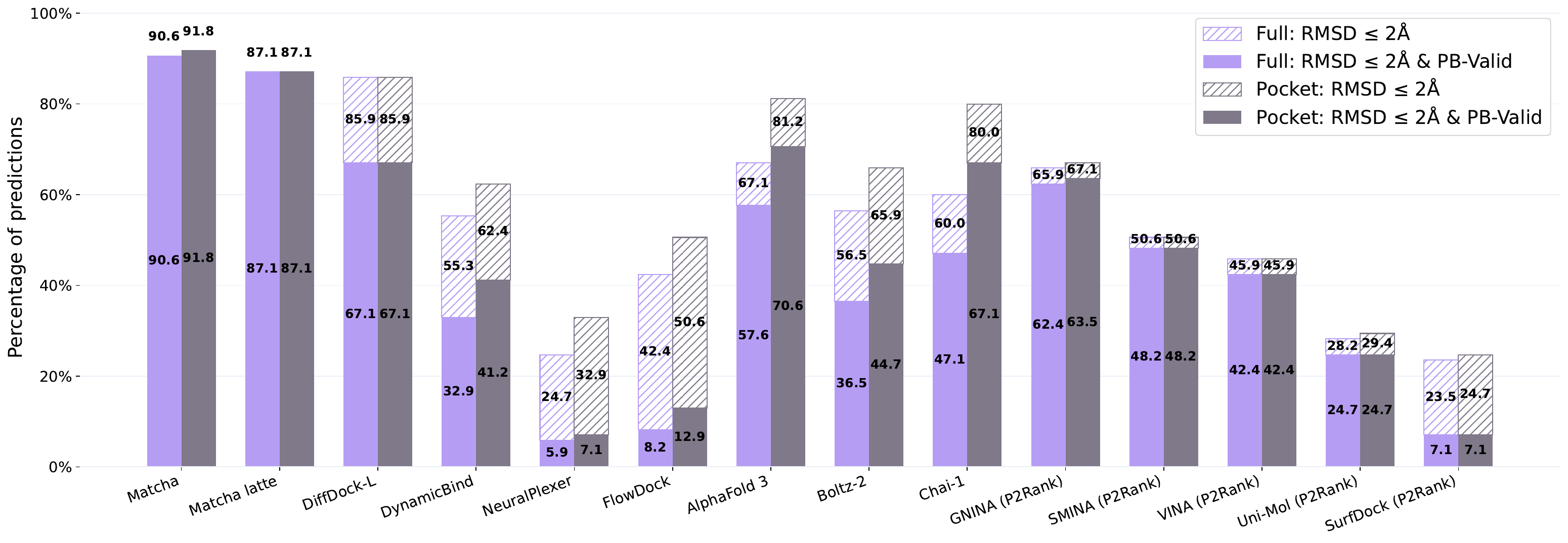}
\caption{Comparison of full and pocket-based alignment strategies in blind docking for \textsc{Astex} Diverse set.}
\label{fig:double_astex}

\end{figure*}

\begin{figure*}[!t]
  \centering
  \includegraphics[width=\textwidth]{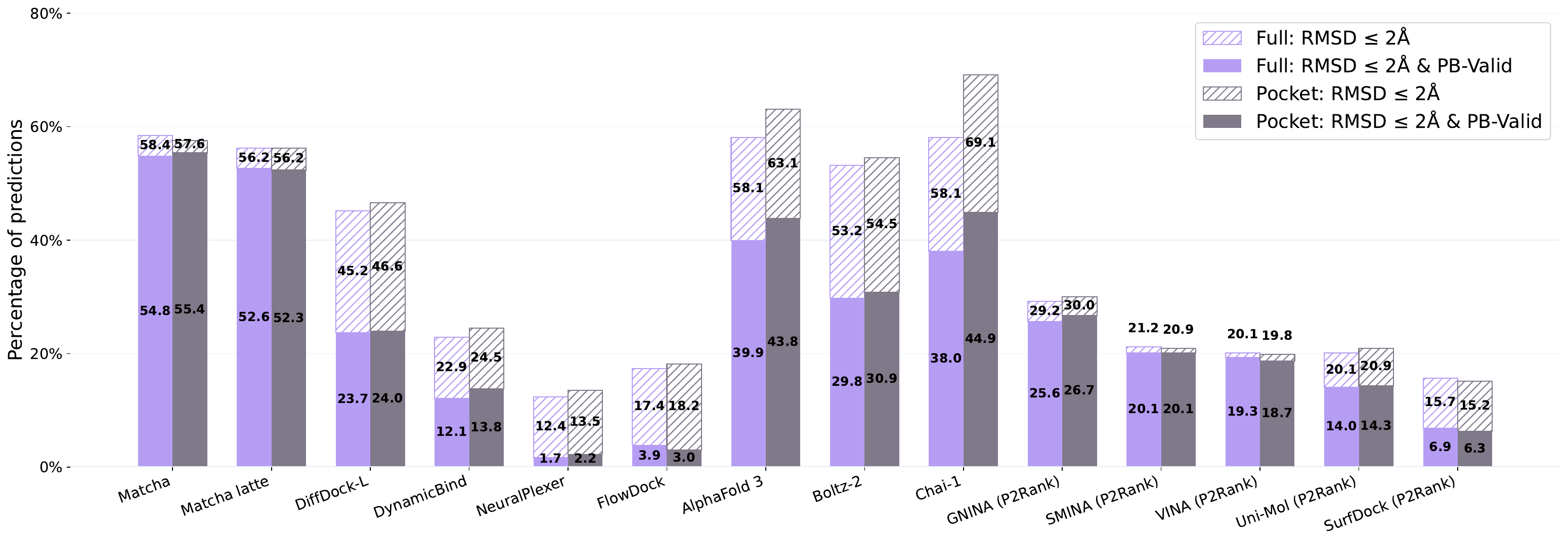}
\caption{Comparison of full and pocket-based alignment strategies in blind docking for \textsc{PDBBind} test set.}
\label{fig:double_pdbbind}

\end{figure*}

\begin{figure*}[!t]
  \centering
  \includegraphics[width=\textwidth]{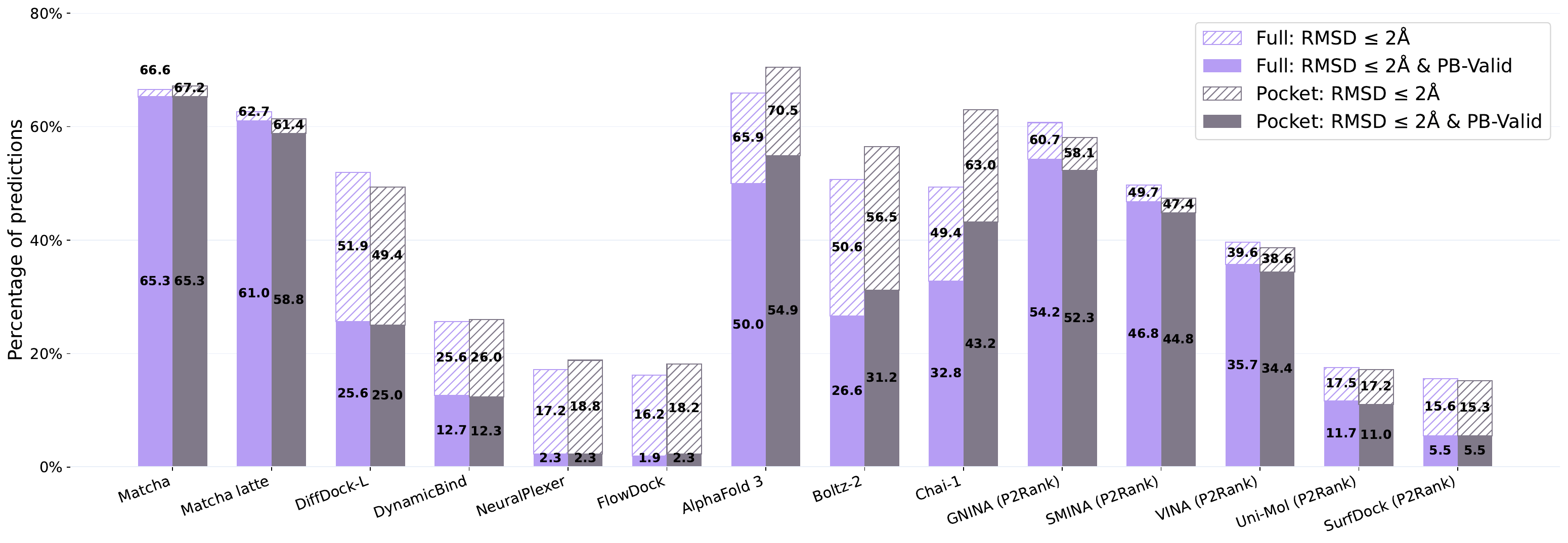}
\caption{Comparison of full and pocket-based alignment strategies in blind docking for \textsc{PoseBusters V2} dataset.}
\label{fig:double_posebusters}

\end{figure*}

\begin{figure*}[!t]
  \centering
  \includegraphics[width=\textwidth]{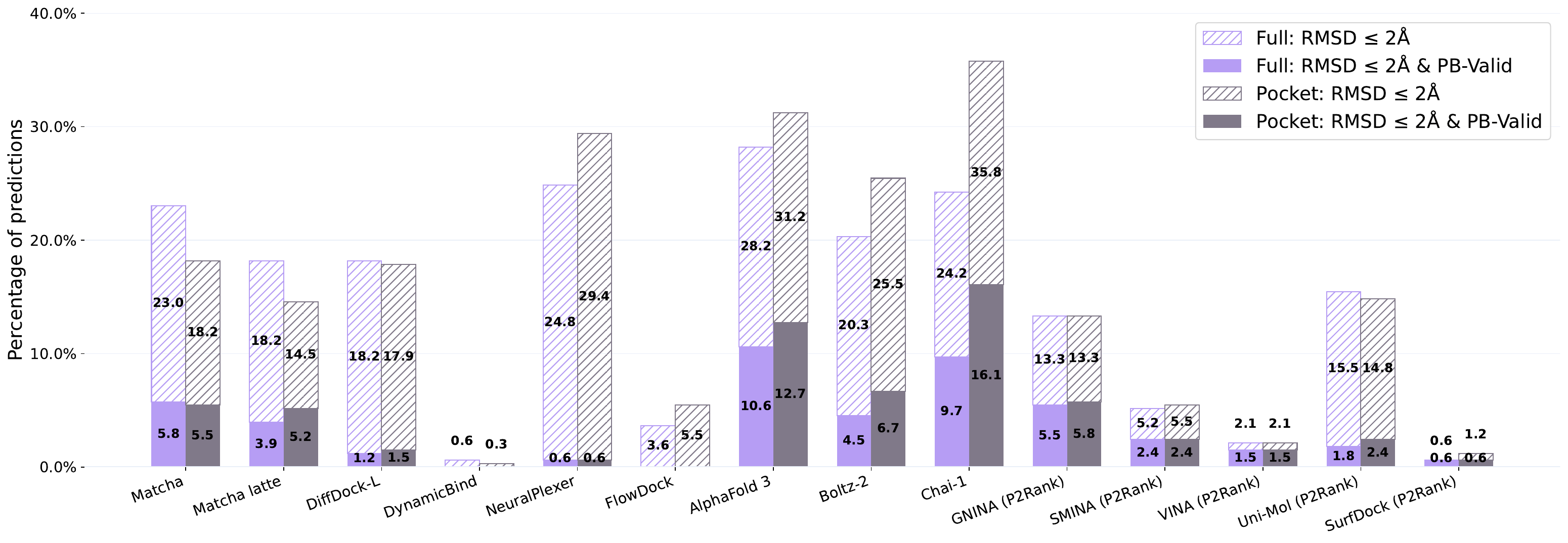}
\caption{Comparison of full and pocket-based alignment strategies in blind docking for \textsc{DockGen} dataset.}
\label{fig:double_dockgen}

\end{figure*}

We computed docking quality metrics using both \textsc{full} and \textsc{pocket-based} approaches for structure prediction methods (\textsc{AlphaFold 3}, \textsc{Boltz-2}, \textsc{Chai-1}, \textsc{NeuralPlexer}, \textsc{FlowDock}, \textsc{DynamicBind}), rigid docking approaches and classical docking methods.
For all four test sets, the obtained results reveal comparable metric inflation across methods that predict protein structure (see Figures~\ref{fig:double_astex}, \ref{fig:double_pdbbind}, \ref{fig:double_posebusters} and \ref{fig:double_dockgen}).
The increase is around 10-20\% in $\mathrm{RMSD}\leq\SI{2}{\angstrom}$ for the \textsc{Astex} dataset and around 5-10\% for \textsc{PoseBusters V2}.
Moreover, the choice of the alignment almost does not affect the ordering of the docking methods: all claims made under \textsc{pocket-based} alignment also hold under \textsc{full} alignment.
Thus, although pocket-based alignment yields higher absolute success rates for methods that predict protein structure, our qualitative conclusions do not depend on this choice.

\section{PoseBusters tests}
\label{app:posebusters_tests}
We report PoseBusters results for the following 27 tests according to the release in \url{https://github.com/maabuu/posebusters/releases/tag/v0.4.5}:
\begin{enumerate}
\item \verb|mol_pred_loaded|,
\item \verb|mol_cond_loaded|,
\item \verb|sanitization|,
\item \verb|inchi_convertible|,
\item \verb|all_atoms_connected|,
\item \verb|bond_lengths|,
\item \verb|bond_angles|,
\item \verb|internal_steric_clash|,
\item \verb|aromatic_ring_flatness|,
\item \verb|non-aromatic_ring_non-flatness|,
\item \verb|double_bond_flatness|,
\item \verb|internal_energy|,
\item \verb|protein-ligand_maximum_distance|,
\item \verb|minimum_distance_to_protein|,
\item \verb|minimum_distance_to_organic_cofactors|,
\item \verb|minimum_distance_to_inorganic_cofactors|,
\item \verb|minimum_distance_to_waters|,
\item \verb|volume_overlap_with_protein|,
\item \verb|volume_overlap_with_organic_cofactors|,
\item \verb|volume_overlap_with_inorganic_cofactors|,
\item \verb|volume_overlap_with_waters|,
\item \verb|double_bond_stereochemistry|,
\item \verb|mol_true_loaded|,
\item \verb|molecular_bonds|,
\item \verb|molecular_formula|,
\item $\mathrm{RMSD}\leq\SI{2}{\angstrom}$,
\item \verb|tetrahedral_chirality|.
\end{enumerate}

\section{Distance-aware attention bias}
\label{app:attention_bias}
This section details the distance-aware attention bias summarized in the Methods.
We adopt the approach from \textsc{UniMol} and \textsc{AlphaFold~3}, where spatial features are used as extra biases in self-attention.
Given 3D coordinates $x=\{x_i\}_{i=1}^N,\; x_i\in\mathbb{R}^3$, we form a per-head attention bias by combining a radial (distance-based) and a directional (vector-based) term. For a pair $(i,j)$ with edge type $t_{ij}\in\{1,\dots,T\}$, define the displacement $\Delta_{ij}=x_i-x_j$ and a stabilized inverse distance
\begin{equation}
s_{ij} \;=\; \frac{1}{\|\Delta_{ij}\|_2^2 + 1}\,.
\end{equation}
An edge-type--specific affine transform produces $\tilde{s}_{ij}=\alpha_{t_{ij}}\, s_{ij} + \beta_{t_{ij}}$. We then embed $\tilde{s}_{ij}$ with a $K$-kernel Gaussian RBF:
\begin{equation}
\phi^{(k)}_{ij} \;=\; \mathcal{N}\!\big(\tilde{s}_{ij}\,;\,\mu_k,\sigma_k^2\big),
\qquad k=1,\dots,K,
\end{equation}
followed by a small MLP projection $g:\mathbb{R}^K\!\to\!\mathbb{R}^H$ to obtain a per-head radial bias $o_{ij}=g(\phi_{ij})\in\mathbb{R}^H$. In parallel, a directional projection $h:\mathbb{R}^3\!\to\!\mathbb{R}^H$ maps the displacement to $v_{ij}=h(\Delta_{ij})$.
The per-pair bias is
\(
b_{ij}=g(\phi_{ij})+h(\Delta_{ij}) \in \mathbb{R}^H.
\)
Stacking over all pairs yields a tensor in $\mathbb{R}^{N\times N\times H}$; we then move the head dimension to the front to obtain
\(
B_{hij} = [b_{ij}]_h \in \mathbb{R}^{H\times N\times N},
\)
which we add to the attention logits.

\section{Training algorithm}
\label{app:training_algorithm}
The general scheme of \textsc{Matcha} docking training is given in Algorithm~\ref{alg:training}.

\begin{algorithm*}[!t]
\caption{General scheme of \textsc{Matcha} docking training}
\label{alg:training}
\begin{algorithmic}[1]
\Require Protein-ligand complexes $D$, initialized flow model $v$,
\Statex \hspace{\algorithmicindent} $\mathrm{stage}\in\{1,2,3\}$, $\sigma_{\mathrm{large}}$, $\sigma_{\mathrm{medium}}$, $\sigma_{\mathrm{small}}$.
\For{each batch of protein-ligand complexes in $D$}
  \State Take the ligand conformation from the batch. Compute its centroid $\mathrm{tr}_{\mathrm{true}}$. Apply the augmentations described in Methods.
  \State Identify the rotatable bonds and let $m$ be their number.
  \State Sample a random translation, rotation and torsion, $\mathrm{tr},\mathrm{rot},\mathrm{tor}$, and apply them to the conformation.
  \State Set $\mathrm{rot}_{\mathrm{true}}:=\mathrm{inverse}(\mathrm{rot})$ and $\mathrm{tor}_{\mathrm{true}}:=\mathrm{inverse}(\mathrm{tor})$.
  \State Sample noise $\mathrm{tr}_{\mathrm{noise}}$, $\mathrm{rot}_{\mathrm{noise}}$, $\mathrm{tor}_{\mathrm{noise}}$:
  \Statex \quad if $\mathrm{stage}=1$:
    \mbox{$\mathrm{tr}_{\mathrm{noise}}\sim\mathcal{N}(0,\sigma^{2}_{\mathrm{large}})$},
    \mbox{$\mathrm{rot}_{\mathrm{noise}}\sim\mathrm{Unif}(\mathrm{SO}(3))$},
    \mbox{$\mathrm{tor}_{\mathrm{noise}}\sim\mathrm{Unif}(\mathrm{SO}(2))^{m}$};
  \Statex \quad if $\mathrm{stage}=2$:
    \mbox{$\mathrm{tr}_{\mathrm{noise}}\sim\mathcal{N}(\mathrm{tr}_{\mathrm{true}},\sigma^{2}_{\mathrm{medium}})$},
    \mbox{$\mathrm{rot}_{\mathrm{noise}}\sim\mathrm{Unif}(\mathrm{SO}(3))$},
    \mbox{$\mathrm{tor}_{\mathrm{noise}}\sim\mathrm{Unif}(\mathrm{SO}(2))^{m}$};
  \Statex \quad if $\mathrm{stage}=3$:
    \mbox{$\mathrm{tr}_{\mathrm{noise}}\sim\mathcal{N}(\mathrm{tr}_{\mathrm{true}},\sigma^{2}_{\mathrm{small}})$},
    \mbox{$\mathrm{rot}_{\mathrm{noise}}\sim\mathcal{N}(\mathrm{rot}_{\mathrm{true}},\sigma^{2}_{\mathrm{small}}\mI)$},
    \mbox{$\mathrm{tor}_{\mathrm{noise}}\sim\mathcal{N}(\mathrm{tor}_{\mathrm{true}},\sigma^{2}_{\mathrm{small}})^{m}$}.
  \State Sample $t\sim\mathrm{Uniform}(0,1)$ and interpolate between the noisy and true transformations.
  \Statex \hspace{\algorithmicindent} Apply the result to the ligand, giving $\vx(t):=(\mathrm{tr}(t),\mathrm{rot}(t),\mathrm{tor}(t))$.
  \State Evaluate the flow model $v(\vx(t),t)\in\R^3\times\mathfrak{so}(3)\times\mathfrak{so}(2)^m$.
  \State Compute the flow-matching loss on each component and take their linear combination.
  \State Take a gradient step on that loss.
\EndFor
\Ensure Trained flow model $v$ for the given stage.
\end{algorithmic}
\end{algorithm*}

\section{Additional benchmark results}
\label{app:additional_results}

\subsection{Robustness on the temporally held-out PoseBusters V2 subset}
\begin{figure*}[!t]
  \centering
  \includegraphics[width=\textwidth]{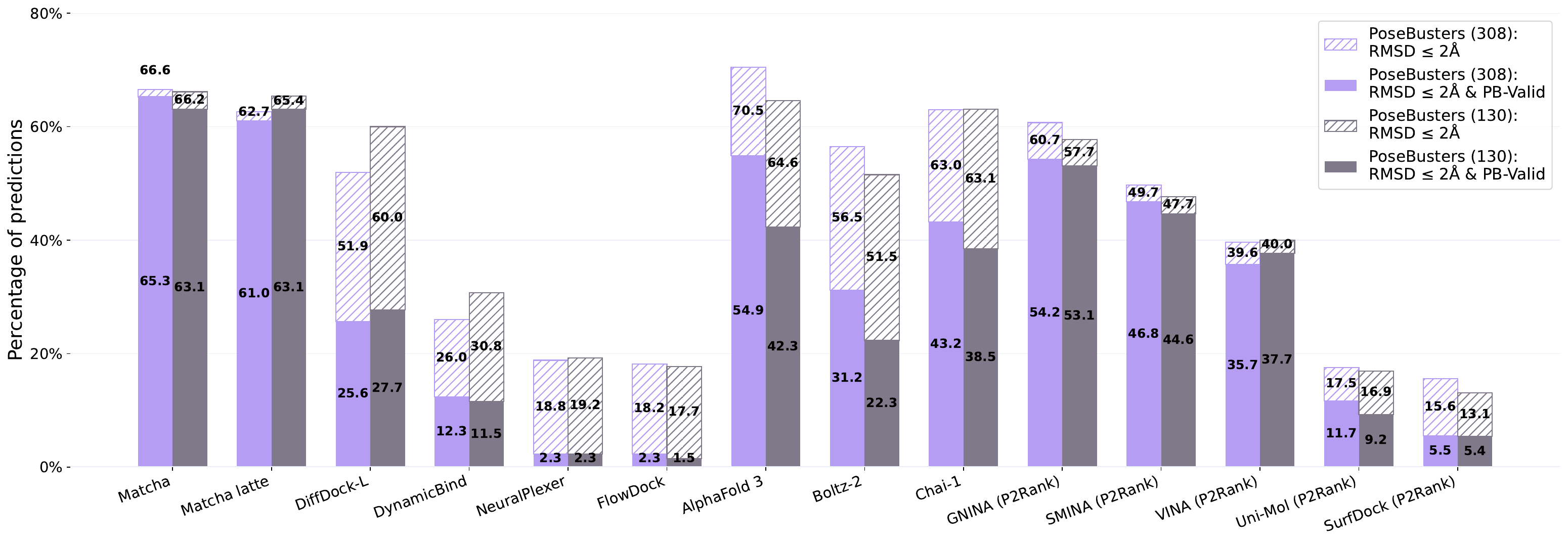}
  \caption{Blind ligand docking success rates on the \textsc{PoseBusters V2} dataset for the full benchmark and the temporally held-out subset.}
  \label{fig:posebusters}
  
\end{figure*}

On the full \textsc{PoseBusters V2} set, co-folding methods report higher numbers; however, a large fraction of those complexes were deposited before the co-folding models' training cutoff (e.g., \textsc{AlphaFold~3} uses a September 2021 cutoff), so the comparison is partly confounded by train--test overlap~\citep{morehead2026assessing}.
On the held-out subset (structures deposited after September 30, 2021), co-folding performance drops (up to $\sim$10\% in $\mathrm{RMSD}\leq\SI{2}{\angstrom}$~\&~PB-valid), while \textsc{Matcha} and other DL-based and classical docking methods remain stable, indicating that \textsc{Matcha}'s strong results on \textsc{PoseBusters V2} are robust under fair evaluation (Fig.~\ref{fig:posebusters}).

\subsection{Number of Riemannian Euler integration steps}
We evaluated the number of Riemannian Euler integration steps used during inference on \textsc{Astex}, \textsc{PDBBind} and \textsc{PoseBusters V2}.
Increasing the number of steps from 1 to 10 improves both $\mathrm{RMSD}\leq\SI{2}{\angstrom}$ and $\mathrm{RMSD}\leq\SI{2}{\angstrom}$~\&~PB-valid success rates.
Beyond 10 steps, the gains saturate, and increasing the budget to 15 steps does not provide a measurable improvement.

\begin{figure*}[!t]
  \centering
  \figurepanelspreamble
  \panelref{0.325\textwidth}{a}{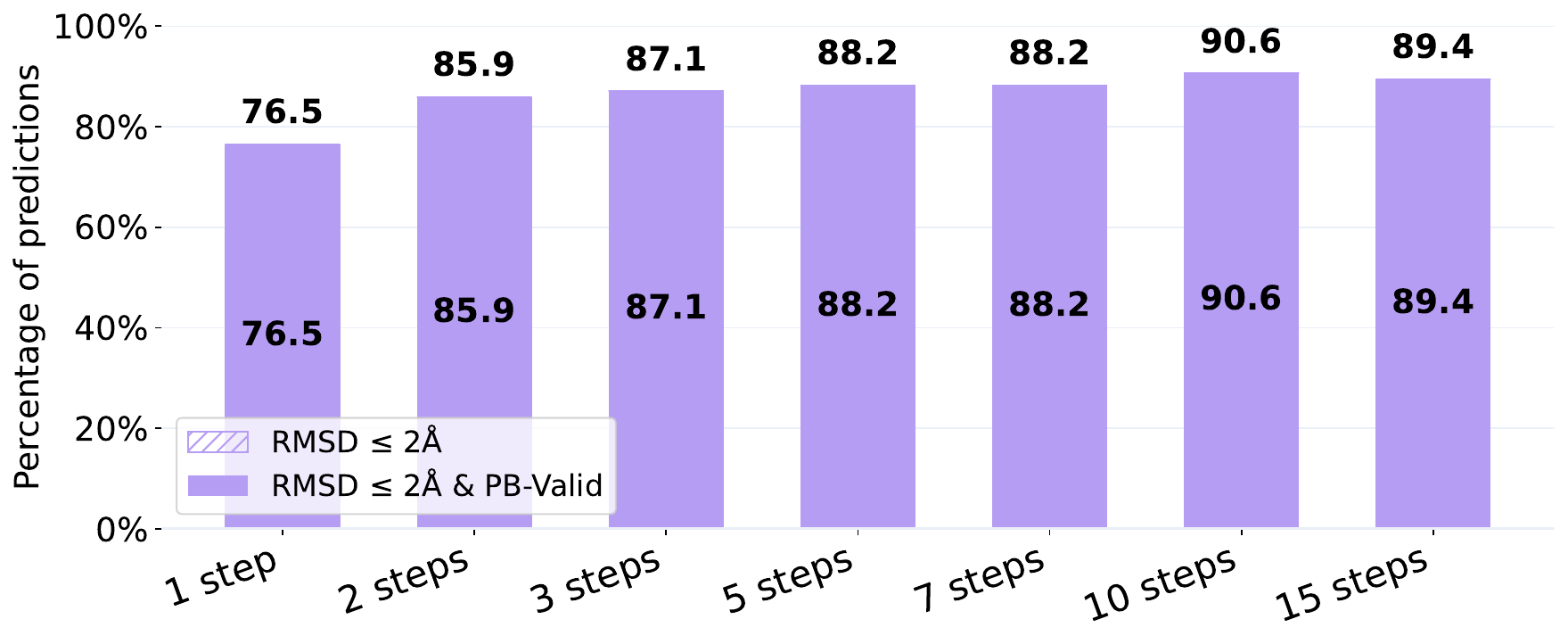}{\label{fig:res_euler_ablation_astex}}
  \hfill
  \panelref{0.325\textwidth}{b}{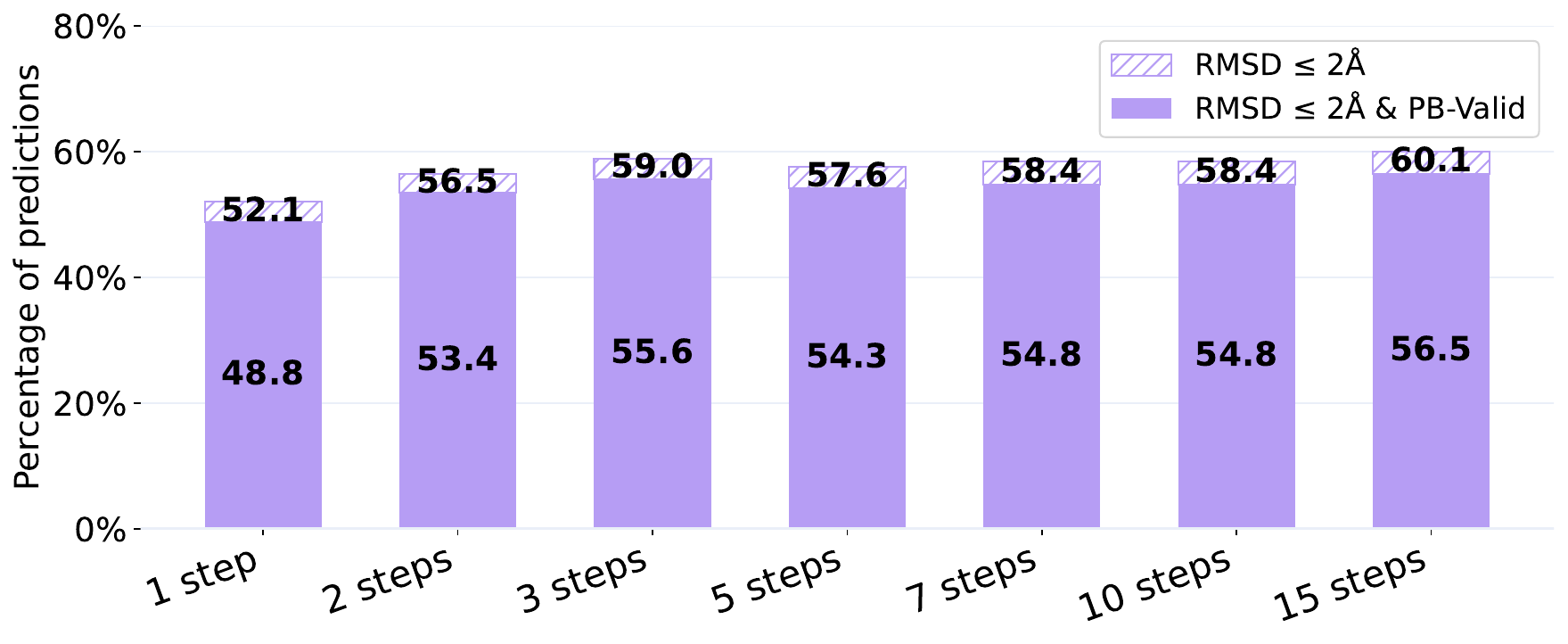}{\label{fig:res_euler_ablation_pdbbind}}
  \hfill
  \panelref{0.325\textwidth}{c}{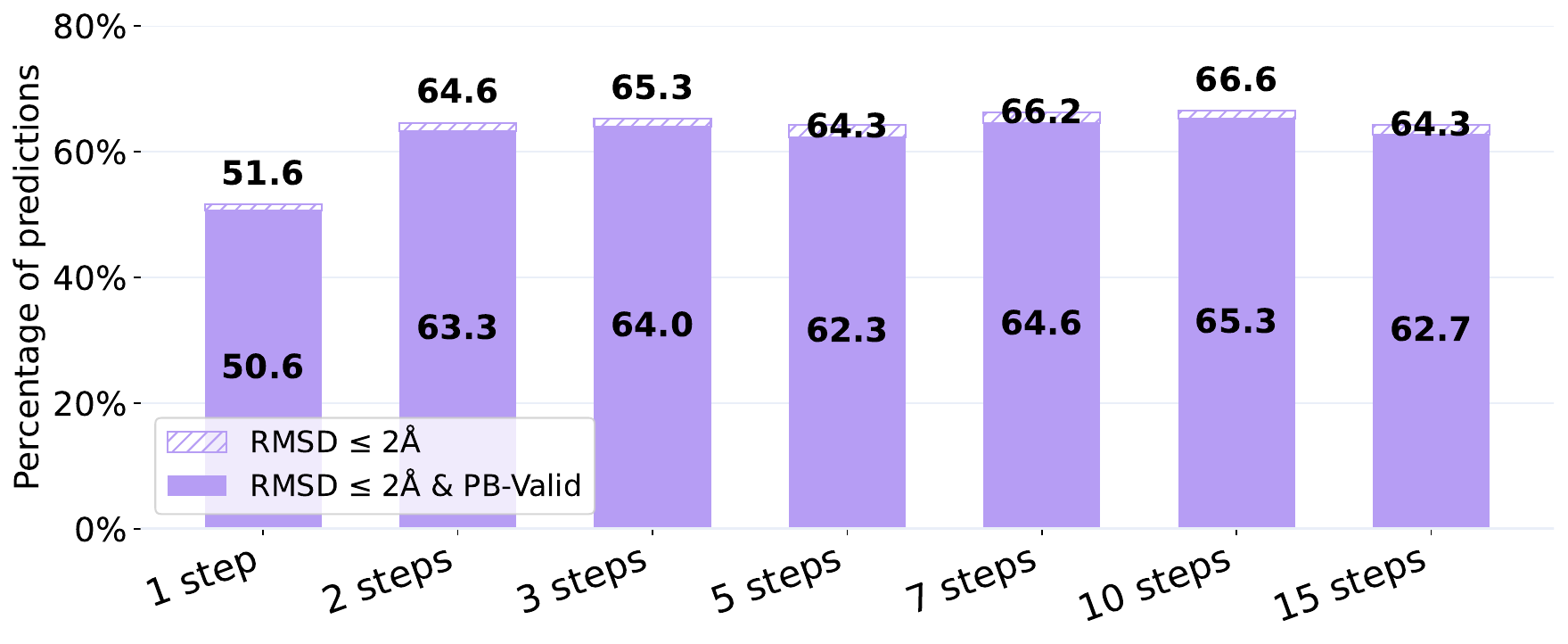}{\label{fig:res_euler_ablation_posebusters}}
  \figurepanelspostamble
  \caption{\textbf{Effect of the number of Riemannian Euler integration steps.}
  We evaluate the inference-time integration budget by varying the number of Euler steps used by \textsc{Matcha}.
  \cappanel{a--c} Blind docking success rates as a function of the number of Euler steps on \textsc{Astex}, \textsc{PDBBind} and \textsc{PoseBusters V2}, respectively.
  Success is reported as the percentage of predictions with symmetry-corrected heavy-atom RMSD $\leq\SI{2}{\angstrom}$ and, where indicated, satisfying the PoseBusters validity criteria.}
  \label{fig:euler_steps_extended}
  
\end{figure*}

\subsection{Matcha post-processing applied to the baseline methods}
\label{app:postprocessing_baselines}

\begin{figure*}[!t]
  \centering
  \figurepanelspreamble
  \panelref{0.49\textwidth}{a}{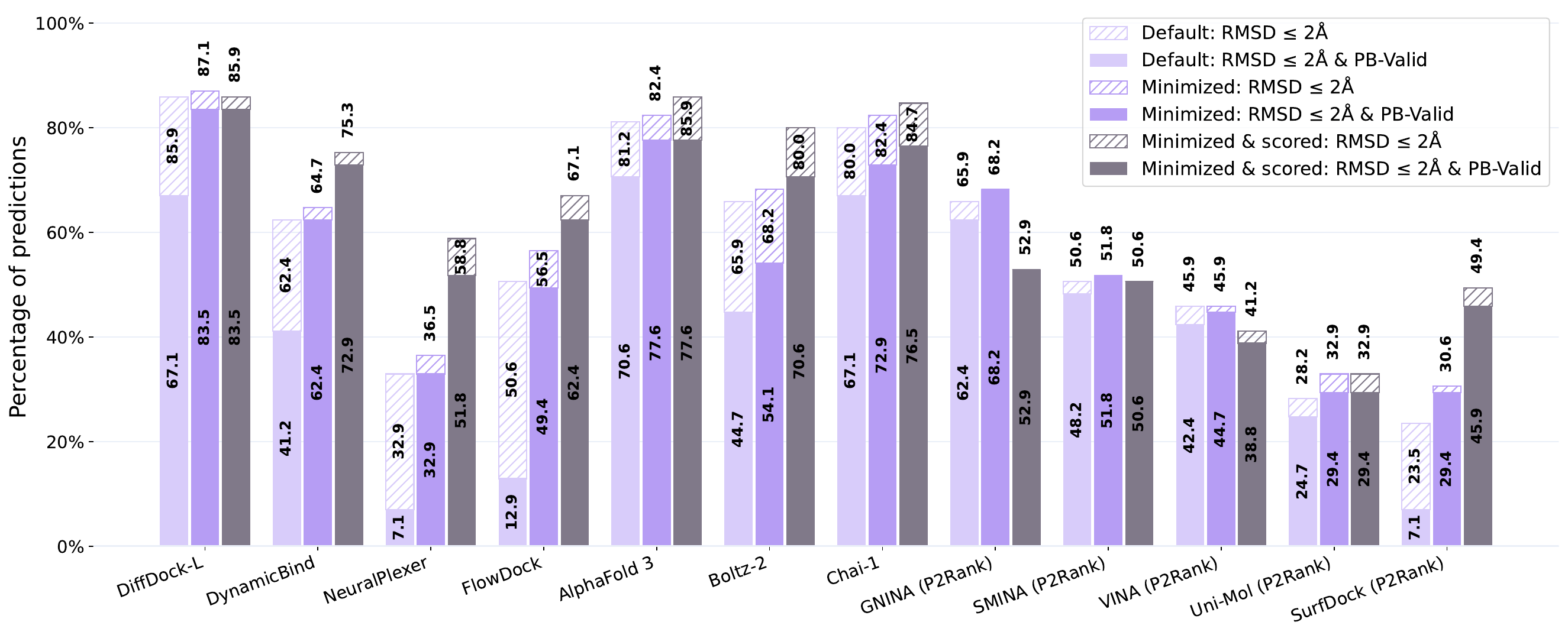}{\label{fig:postproc_baselines_astex}}
  \hfill
  \panelref{0.49\textwidth}{b}{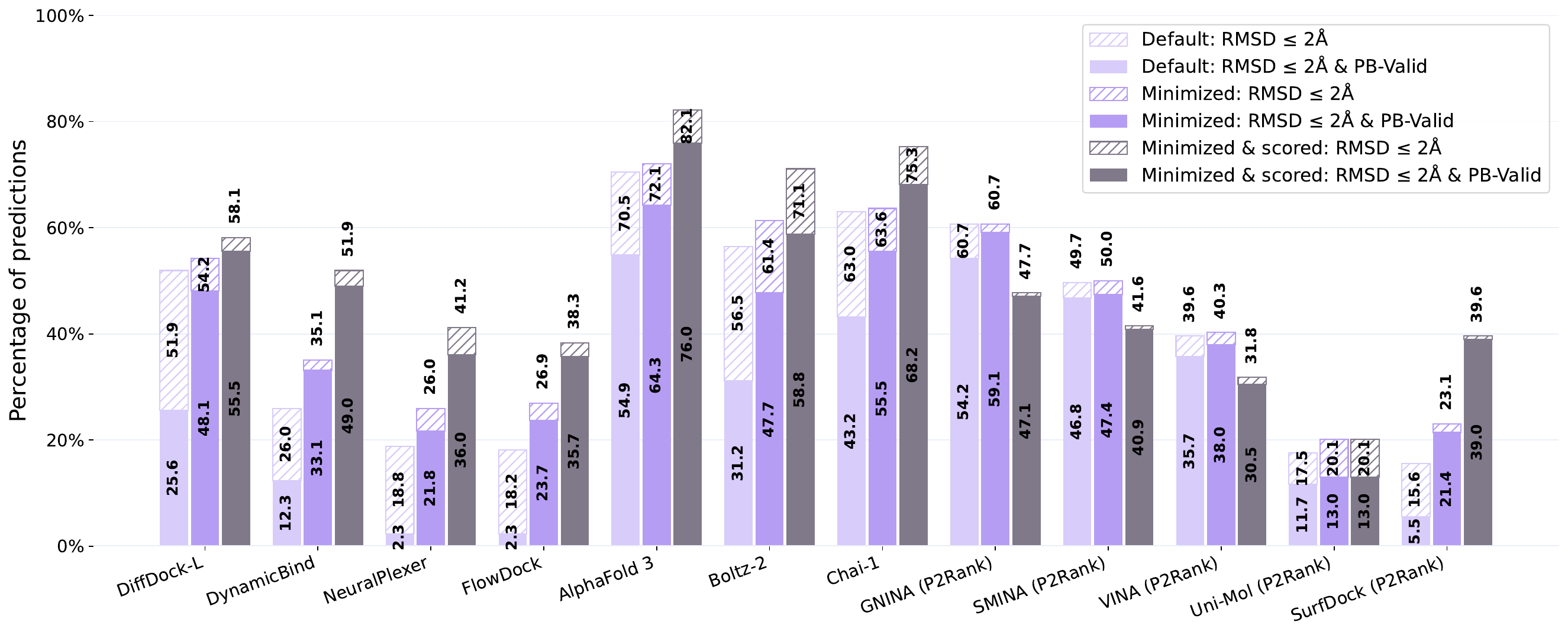}{\label{fig:postproc_baselines_posebusters}}

  \vspace{0.6em}

  \panelref{0.49\textwidth}{c}{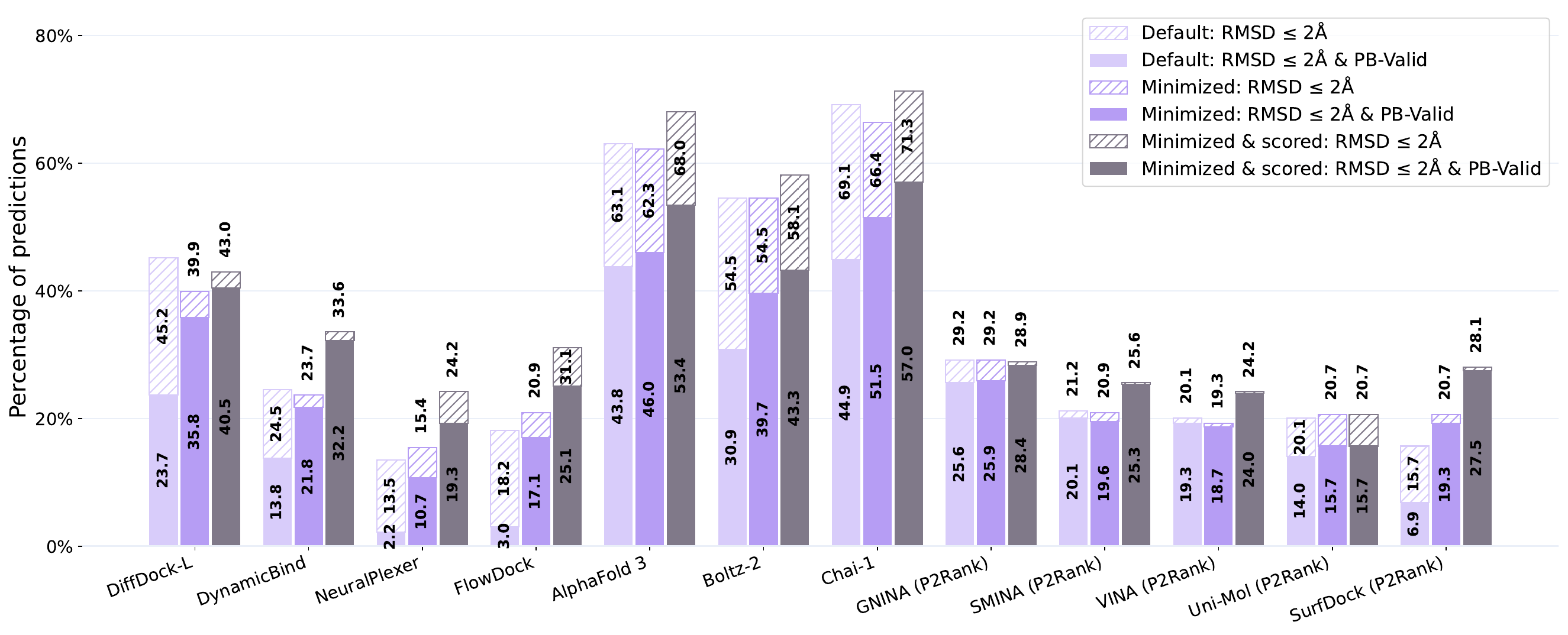}{\label{fig:postproc_baselines_pdbbind}}
  \hfill
  \panelref{0.49\textwidth}{d}{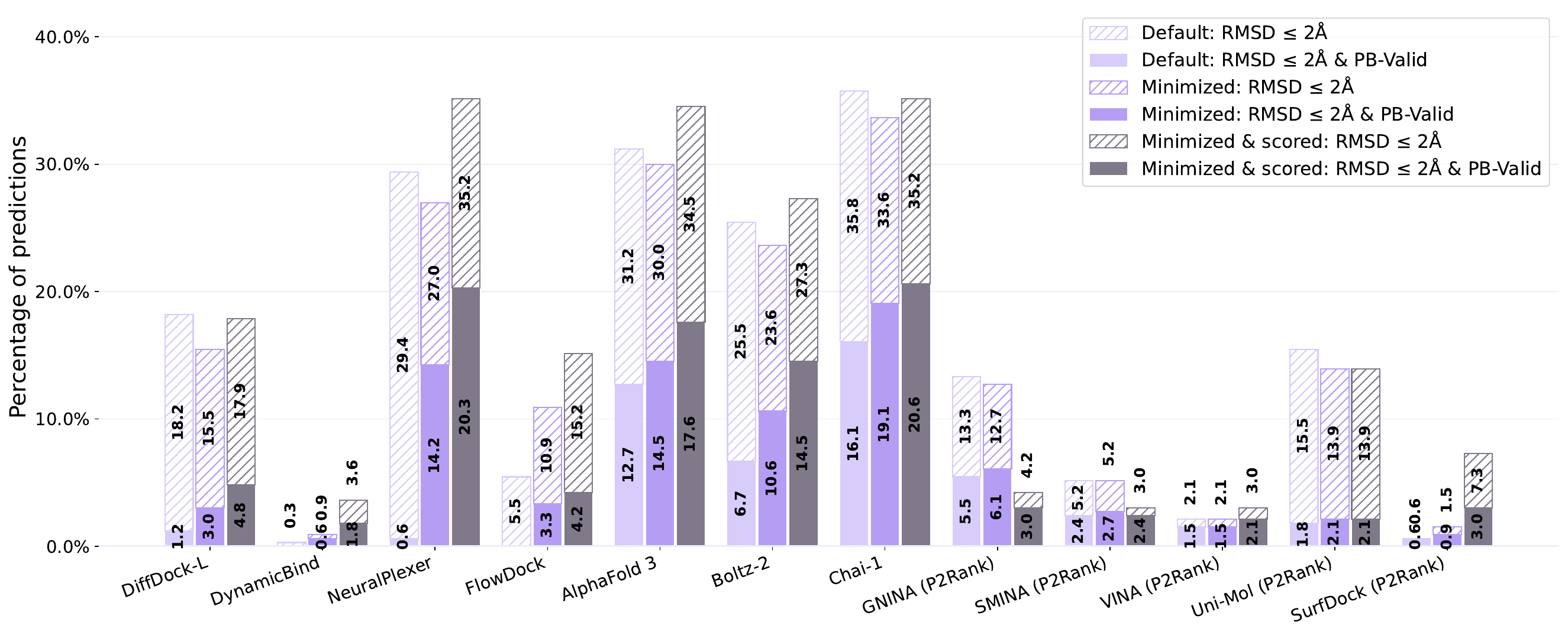}{\label{fig:postproc_baselines_dockgen}}
  \figurepanelspostamble
  \caption{Effect of the \textsc{Matcha} post-processing module applied to the baseline methods.
  \cappanel{a} \textsc{Astex}.
  \cappanel{b} \textsc{PoseBusters V2}.
  \cappanel{c} \textsc{PDBBind}.
  \cappanel{d} \textsc{DockGen}.}
  \label{fig:postproc_baselines}
  
\end{figure*}

The \textsc{Matcha} post-processing module is method-agnostic: it takes candidate poses, minimizes them with \textsc{GNINA}, removes physically implausible complexes with the validity filters of Appendix~\ref{app:posebusters_tests}, and ranks the survivors with the default scoring rule (Eq.~\ref{eq:gnina_scoring}). To separate the contribution of the flow-matching generator from the contribution of post-processing, we applied exactly the same module to the poses produced by every baseline method on all four benchmarks (Figs.~\ref{fig:postproc_baselines_astex}--\ref{fig:postproc_baselines_dockgen}). For each method we report three settings: \emph{Default}, the best pose is selected by the method; \emph{Minimized}, the best method's pose is minimized with \textsc{GNINA}; and \emph{Minimized \& scored}, the best pose is selected by the default post-processing pipeline of \textsc{Matcha} (including minimization, validity filtering and \textsc{GNINA}-based scoring).

\begin{figure*}[!t]
  \centering
  \includegraphics[width=0.85\textwidth]{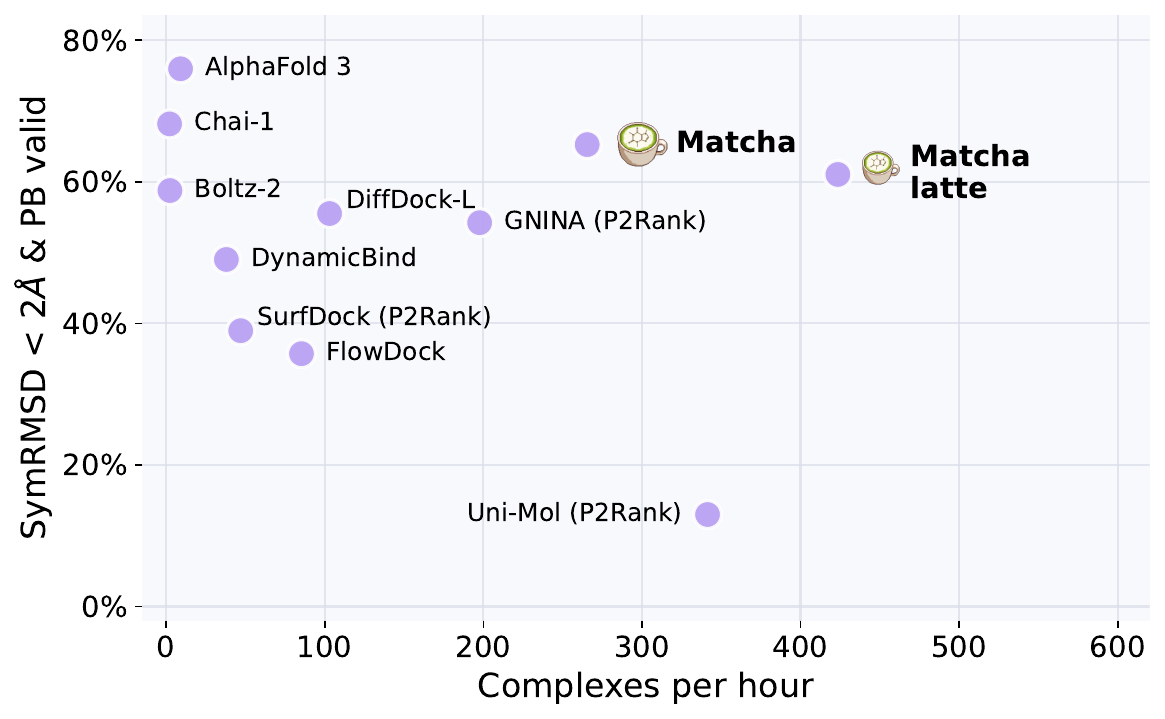}
  \caption{Speed--accuracy trade-off on \textsc{PoseBusters V2} when every method receives the same \textsc{Matcha} post-processing module (\emph{Minimized \& scored} success rate; see Figs.~\ref{fig:postproc_baselines_astex}--\ref{fig:postproc_baselines_dockgen}). Complexes per hour is computed on one NVIDIA A100 40GB GPU.}
  \label{fig:timing_with_postproc}
  
\end{figure*}

Post-processing improves every method, and the improvement is largest for methods whose raw output violates physical constraints. On \textsc{Astex}, $\mathrm{RMSD}\leq\SI{2}{\angstrom}$~\&~PB-valid success rises from 67.1\% to 83.5\% for \textsc{DiffDock-L} and from 70.6\% to 77.6\% for \textsc{AlphaFold~3}, compared with $89.2\pm1.5$\% for \textsc{Matcha}. On \textsc{PoseBusters V2} the same procedure lifts \textsc{AlphaFold~3} from 54.9\% to 76.0\% and \textsc{Chai-1} from 43.2\% to 68.2\%, above the 65.8\% reached by \textsc{Matcha}; on \textsc{PDBBind}, post-processed \textsc{Chai-1} reaches 57.0\% versus 54.2\% for \textsc{Matcha}. 
At the same time, classical docking approaches do not benefit from the additional minimization and re-scoring of poses since their original pipelines already do it.
We therefore do not claim that \textsc{Matcha} dominates co-folding models once they are given the same physical refinement: the advantage of \textsc{Matcha} is that it attains comparable or better physically valid accuracy at a small fraction of the computational cost. Fig.~\ref{fig:timing_with_postproc} summarizes this trade-off on \textsc{PoseBusters V2}: success uses the \emph{Minimized \& scored} setting (the same \textsc{Matcha} post-processing module applied to every method) except for \textsc{GNINA} which already minimizes its poses itself, and updated throughput is reported as complexes per hour on one NVIDIA A100 40GB GPU, adding this shared post-processing time to each method's own inference time. The timing figure in the main Results uses each method's default pipeline without this shared post-processing step.

\end{appendices}

\end{document}